\documentclass[11pt]{report}
\usepackage{comment}
\usepackage{pgfplots}
\usepackage{tikz}
\pgfplotsset{compat = newest}
\usepackage{times}
\usepackage{helvet}
\usepackage{courier}
\usepackage{setspace}
\usepackage{amsmath}
\usepackage{layout}
\usepackage{latexsym}
\usepackage{amssymb}
\usepackage{url}
\usepackage{hyperref}
\usepackage[export]{adjustbox}
\usepackage[amsmath,hyperref,thmmarks]{ntheorem}
\usepackage{color}
\usepackage{alltt}
\usepackage{float}
\usepackage{afterpage}
\usepackage{longtable}
\usepackage{graphicx}
\usepackage{relsize}

\begin{document}

\renewcommand{\mp}{\mbox{modus ponens}}
\newcommand{\msp}{\mbox{$\;\:$}}
\newcommand{\mpsp}{\mbox{modus ponens} }
\newcommand{\smtt}[1]{\mbox{\small\tt #1}}
\newcommand{\mtt}[1]{\mbox{\tt #1}}
\newcommand{\fmtt}[1]{\mbox{\footnotesize\tt #1}}
\newcommand{\mbf}[1]{\mbox{\bf #1}}
\newcommand{\temv}[1]{\mbox{\em #1\/}}
\newcommand{\ben}{\begin{enumerate}}
\newcommand{\een}{\end{enumerate}}
\newcommand{\bit}{\begin{itemize}}
\newcommand{\eit}{\end{itemize}}
\newcommand{\bv}{\begin{verbatim}}
\newcommand{\ev}{\end{verbatim}}
\newcommand{\bdesc}{\begin{description}}
\newcommand{\edesc}{\end{description}}
\newcommand{\iensp}{\mbox{i.e.}}
\newcommand{\egnsp}{\mbox{e.g.}}
\newcommand{\g}{\mbox{GPT-4}}
\newcommand{\mdot}{\mbox{$\;$.$\;$}}
\newcommand{\sdot}{\mbox{$\,$.$\,$}}
\newcommand{\miff}{\mbox{\tt$\:$<==>$\:$}}
\newcommand{\gsp}{\mbox{GPT-4} }
\newcommand{\gnsp}{\mbox{GPT-4}}
\newcommand{\bigtilde}{\raisebox{0.12\baselineskip}{\scalebox{0.8}{$\sim\:$}}}
\newcommand{\mtilde}{\raisebox{0.12\baselineskip}{\scalebox{0.55}{$\sim$}}}

\title{{\bf {\Large GPT-4 Can't Reason}} \\[0.1in] {\small (Position Paper)}}
\author{Konstantine Arkoudas \\[0.05in] {\small Dyania Health}}
\maketitle

\section*{Abstract}
\gsp was released in March 2023 to wide acclaim, marking a very substantial improvement across the board  
over GPT-3.5 (OpenAI's previously best model, which had powered the initial release of ChatGPT). 
Despite the genuinely impressive improvement, however, there are good reasons to be highly skeptical 
of \gnsp's ability to reason. This position paper discusses the nature of reasoning; criticizes the current 
formulation of reasoning problems in the NLP community and the way in which the reasoning performance of LLMs 
is currently evaluated; introduces a collection of 21 diverse reasoning problems; and performs a detailed 
qualitative analysis of \gnsp's performance on these problems. Based on the results of that analysis, 
this paper argues that, despite the occasional flashes of analytical brilliance, \gsp at present is utterly incapable of reasoning.

\section{Introduction}

In early January I wrote 
\href{https://medium.com/@konstantine_45825/chatgpt-is-no-stochastic-parrot-but-it-also-claims-that-1-is-greater-than-1-e3cd1fc303e0}{a 
commentary}\footnote{A modified version of that is being published in the journal {\em Philosophy \& Technology}.}  
presenting an informal evaluation of ChatGPT
across a broad range of subject areas: conventional NLU, folk physics, information retrieval, 
pragmatics, theory of mind, spatial inference, simple logical reasoning, and math. 
The key takeaways were that ChatGPT was a seminal breakthrough; that LLM-based systems
are not mere stochastic parrots but build genuine abstractions and can exhibit creativity; 
that such systems will enable a large array of new and exciting applications;
and that, despite all of the above, these systems are still severely limited when 
it comes to reasoning. 

\gsp was released a couple of months after that,  delivering very substantial improvements across the board. 
I remain impressed and excited by the general capabilities and potential of LLMs, 
and I have little doubt that their performance will continue to improve in the near future.
Nevertheless, there are increasing grounds for skepticism concerning their reasoning abilities. 
In this position paper I will argue  that the best LLM at this time, \gnsp, is  
utterly incapable of reasoning, in spite of its sporadic displays of ingenuity. 

I will largely steer clear of the much broader---and more vague---debate about whether LLMs {\em in general\/}
are capable of (consistently robust) reasoning, but a few brief remarks will help to set the stage and clarify 
why it makes sense to restrict attention to a specific LLM. 
On one side of that broader debate, rosy predictions 
by LLM enthusiasts rely excessively on 
ever-changing scaling ``laws'' that rest on flimsy empirical evidence and on a host of questionable modeling assumptions, 
ill-understood concepts (such as ``emergent'' LLM properties\footnote{The notion 
of an emergent property is clear enough, at least at a high enough level. What is not clear is the relationship between such 
properties and LLM architectures, their basic configurations (number of parameters, compute budget,
dataset size, and so on), and more importantly, important tasks such as reasoning.}), and 
a somewhat dogmatic belief that minimizing cross-entropy loss on next-token prediction
over a huge corpus will deliver a general reasoning engine 
via the magic of transfer learning and the construction of generic 
higher-level representations. 

On the other side of the debate, 
while LLM skeptics have serious arguments to make, those 
arguments are mostly a priori and somewhat vague (for instance, that 
LLMs lack ``a model of the world''), and I do not think they settle the question.
In my view, the most compelling a priori considerations against the plausibility of reliably robust 
LLM reasoning turn on computational complexity results. Reasoning is a (very) computationally hard problem. 
In fact, in  the general case (first-order or higher-order logic), it is algorithmically undecidable,
\iensp, every bit as unsolvable as the halting problem. Thus, by Church's thesis, we cannot expect 
{\em any\/} algorithm, LLMs included, to solve arbitrary reasoning problems in a sound and complete
way.\footnote{Or with perfect precision and recall, to put it---more loosely---in ML-like terms.} 
But even ``easier'' classes of reasoning problems\footnote{Of which there are many:
propositional logic, the two-variable fragment of first-order logic, the Ackerman fragment, 
the guarded fragment, various quantifier-prefix fragments, and so on.} 
typically have either exponential or at least nontrivial polynomial-time complexity profiles.
Problem classes that have linear-time inference algorithms, such as Horn clauses over literals,
are rarely expressive enough.  This tradeoff between generality and expressivity on the one hand
and tractability on the other means that no LLM, no matter how large or how extensively
and cleverly trained and tuned, will ever be able to crack an {\em arbitrary\/} reasoning problem. 
And this is consistent with the famous ``no free lunch'' theorem of machine learning, which points to 
a similar inverse relationship between model generality and performance. 

But LLM advocates can make a couple of cogent counterpoints, while granting that there will 
never be an AI oracle that can essentially solve the halting problem. 
First, they can point out that even though a problem 
might have high worst-case asymptotic complexity, it might still be solvable well enough 
{\em in practice}. Unlike random instances, real-world instances of reasoning problems 
(and indeed real-world instances of most computationally hard problems) appear to have 
structure that allows clever algorithms to tackle them effectively.\footnote{Understanding that
structure and rigorously characterizing its relationship with algorithm performance 
(e.g., via different problem parameterizations, such as clause/variable ratios in the case of SAT) 
is a key open problem in theoretical computer science, but that is another matter.}
There are many examples here, from the simplex algorithm for linear programming
and SAT solvers to term unification algorithms and even automatic theorem provers for 
full first-order logic. All of these problems are hard (having at least exponential-time worst-case complexity), 
yet somehow we have algorithms for them that seem to work successfully on a wide variety of inputs. 

Second, and perhaps more important, we need not aim for an oracle anyway. Humans are not oracles either, 
nor do they seem to follow any particular algorithm that captures any one specific class of 
reasoning problems. The ability of humans to reason is much more fluid and messy, but impressive
nevertheless. Is it impossible to build something like an LLM-based system with the reasoning
ability of a well-trained engineer of average intelligence (which perhaps can then become even more intelligent and better 
trained by an endless process of learning and improvement)?

I don't think that building such a system can be ruled out on a priori grounds 
(and here I differ from hard-core AI skeptics). I think it's implausible, for a number 
of reasons,\footnote{Humans do not seem to solve problems by predicting the most likely sequence of tokens to generate. 
They think, explore, experiment, engage in protracted conversation with the people who posed
the problem (sometimes over weeks, months, or even years), refine, generalize, come up with new concepts and terminology, prove results, 
make and refute conjectures, apply heuristics, execute algorithms, analyze and synthesize, and iterate. 
But \temv{how} solutions are generated is one thing and \temv{what}
solutions are generated is another, and that's why it's not incoherent to speak of a model 
whose reasoning performance is roughly at the same level as that of an average human engineer. 
Such a claim can be understood operationally, to mean that a given LLM is able to produce roughly 
the same solutions that we might reasonably expect an average human engineer to produce
(though obviously on a very different time scale).}
but ultimately this strikes me as an empirical question that must be decided on
a case-by-case basis, by subjecting a specific system to testing, \iensp, by interrogating it, 
probing it, and analyzing its responses.  And the case I will consider here is 
that of \gnsp, which appears, by all accounts, to be the most capable LLM at present. 

There are two questions that must be addressed before we proceed.
First, we must agree on what reasoning is, and second, we must say something 
about methodology. The next section contains a brief discussion of reasoning,
but for those who wish to skip that section and dive right into the problems, 
the upshot is that we'll focus on (a liberal conception of) deductive reasoning.
Regarding methodology, just like the January piece, my evaluation here is not based 
on a corpus or set of corpora. Instead, I present a detailed qualitative analysis of \gnsp's 
performance on 21 simple reasoning problems across a wide range of areas,  
most of which have been made up from scratch, while the rest (such as 
Wason's selection task) have been manually tweaked so as to make them less recognizable 
to the model. 

This is done partly to avoid data contamination, which is a serious problem affecting corpus-based evaluations. 
Given how little we know about the training regimen of ChatGPT, 
it is impossible to know for sure whether any existing dataset or problem has effectively been ``seen'' by 
the model during its pretraining or subsequent alignment, whether we're talking about NLP datasets, medical licensing 
exams, Python programming problems, LSAT or bar-entrance exams, SAT or GRE tests, and 
so on.\footnote{According to the analysis carried out by the \href{https://hitz-zentroa.github.io/lm-contamination/}{\fmtt{lm-contamination index}}, 
well-known NLP datasets such as Squad, CoNLL03, MNLI, and others, are indeed contaminated, while several
 others are at best suspicious.}
The qualification ``effectively'' is important, because even though a specific problem might not have been 
seen in its {\em exact\/} form (in a string-matching sense), an essentially equivalent 
variant with a different surface formulation might well have been. 
Hence, simple contamination tests based on substring checks, such as those carried out 
by OpenAI in their \href{https://arxiv.org/abs/2303.08774}{GPT-4 Technical Report} \cite{openai2023gpt4} (posted in March 2023), 
are not sufficient to guarantee lack of contamination.\footnote{In fact, the substring checks carried out by OpenAI 
were not even applied on the entire problem instance, only on 3 randomly selected substrings of 50 
characters each. This is not enough to ensure disjointness for long (or even moderately long) problems,
which are quite common in tests like the UBE (Uniform Bar Exam).}

The absence of a large corpus makes the discussion more qualitative rather than quantitative. 
However, the results are arguably more informative than a numeric metric
computed over a corpus, for a number of reasons. First, because contamination can be ruled out conclusively;
second, because the problems span a large gamut of areas; and third, because a qualitative
discussion of a problem allows for greater depth of analysis and more context in which 
to interpret the results. By contrast, the only way to perform 
a truly informative quantitative evaluation is to come up with a brand new corpus 
that satisfies all of the following criteria: (a) originality; (b) uniformly high quality; 
(c) sufficiently large size; and (d) diversity (not being limited to one type of task only). 
This is a very challenging undertaking.
Even then, a few simple numeric metrics on a brand new dataset might not be particularly illuminating. 
Are the numbers measuring the right things? Do we even know the right things
to measure? Is there an appropriate backdrop in which the numbers can be understood?
For deeper insight, we need to put individual examples under a magnifying glass. 

This is particularly important because we need to scrutinize 
the explanations (``chains of thought'') generated by a reasoner. Unfortunately, almost all 
reasoning corpora comprise either multiple-choice questions or binary classification problems (e.g., ``Does 
sentence $p_2$ follow from premise $p_1$, yes or no?''). Why? Mostly because it is easy to mechanically
evaluate model performance on such datasets. But even in the absence of contamination,
this type of test set runs the serious risk that the LLM will manage to pick the right answers
by latching on to spurious statistical regularities, \iensp, to arrive at the right answers for
the wrong reasons \cite{McCoy2019,niven-kao-2019-probing}.\footnote{Models have been shown to 
leverage the presence of certain cue words (especially negation words) and to 
formulate quick-and-dirty (\iensp, unsound) heuristics such as lexical overlap, 
subsequence, and constituency \cite{McCoy2019}. Most of these results are from 2019 
and revolve around BERT, but more recent work \cite{wang2023robustness} has shown 
that while larger foundational models such as ChatGPT are more robust to input 
perturbations and OOD (out-of-distribution) samples, these continue to be challenges,
suggesting that even ChatGPT-scale models learn unsound shortcuts.} Adversarial augmentation 
of an existing dataset might help, especially if we know what we are trying to guard against, 
but unless an adversarial version restores near-random performance, this can 
quickly devolve into a game of whac-a-mole, where we detect a new round of bogus regularities
exploited by the model and must undertake a new round of adversarial interventions. 

Ultimately, there is really no proper way to assess the reasoning
ability of a system unless we ask it to explain its output. This is an essential part of reasoning,
which is not about producing the right answer by hook or by crook but about {\em deriving\/} the 
right answer {\em for the right reasons}. And rote metrics like ROUGE-L are not fit for purpose here. 
We need to roll up our sleeves and analyze LLM explanations and proof attempts manually. 
We also need to gauge their performance in a dialog setting (e.g., what happens when 
a reasoning error is pointed out to them?). This is the sort of analysis undertaken 
in this paper. I believe the results show unequivocally that \gsp cannot reason. 
The errors are too pervasive and too egregious. \gsp doesn't solve even one of 
the 21 problems discussed here. But much more concerning are the fundamentally flawed
explanations and proof attempts it produces along the way. 

LLM believers will probably demur: {\em But humans also make mistakes, 
and surely we're not prepared to say that humans can't reason just because they make mistakes}? 
First, it is not accurate to say without qualification that ``humans can reason,'' 
certainly not in the sense 
that we can randomly pluck any person from the street and expect them to reliably perform 
normatively correct reasoning. 
Most neurobiologically normal humans have the \temv{capacity} to become proficient in reasoning, 
but actually attaining such proficiency takes significant training and discipline. Humans are known
to be susceptible to a large assortment of cognitive biases, which can only be overcome by rigorous 
instruction. Focusing on the reasoning skills of untrained people is a bit like focusing on 
the singing skills of the general population. Everybody sings in the shower, but without formal 
training (or at least exceptional talent) the results are usually regrettable. 

Of course, even sophisticated human reasoners make mistakes, just like trained singers can hit false notes. 
But if a human made \temv{these} mistakes, the ones reported in this article,
then I would conclude without any hesitation that they cannot reason. 
Even if they went on to list a large number of other examples demonstrating impeccable reasoning, 
I would suspect that other factors (such as rote memorization or cheating) were behind the performance discrepancy. 
For the mistakes reported here are not performance mistakes, 
the sort of innocuous errors that humans might make---and promptly correct---when they are careless or tired. 
If a human made these mistakes, and made them consistently under repeated questioning, 
that would indicate without doubt that they don't have the necessary logical {\em competence},
that they lack fundamental concepts that are part and parcel of the fabric of reasoning, such as logical 
entailment and set membership. And I would certainly not entrust that 
person with generating reams of Python or Javascript code for an enterprise. Nor would I start organizing 
international conferences to investigate how their reasoning prowess might threaten humanity with extinction. 

\section{What is Reasoning?}

Reasoning is not quite the same thing as intelligence, but it's a necessary ingredient for it. 
Broadly put, reasoning is the process of drawing and evaluating {\em conclusions\/} from a given body of information. 
More precisely, it is the process of making and---more importantly---{\em justifying\/} arguments. An argument consists of 
a conclusion (the argument's upshot, so to speak) and a set of {\em premises\/} from which
the conclusion is derived. Premises represent information that
is taken as given, if only provisionally, for the purposes of the argument. The conclusion and the premises 
are typically declarative sentences (expressed either in natural language or in the notation
of a symbolic logic) that can be true or false, but they may also be represented by alternative notational devices, such as diagrams.
We say that a set of premises $S$ logically {\em entails\/} (or logically {\em implies}) a conclusion $p$ iff 
$p$ is true whenever all the sentences in $S$ are true, in which case the argument is said to be {\em valid}. 
This means that it's logically impossible to have a state of affairs in which every element of $S$ holds but $p$ does not. 
This key logical relationship is a linchpin of human reasoning.\footnote{Here we understood premises and conclusions
as syntactic objects (sentences or diagrams), but there are alternative approaches. 
For instance, a semanticist might think of premises and 
conclusions as {\em propositions}, abstract objects capable of being true or false. A sentence 
then {\em expresses\/} or {\em represents\/} a proposition. Propositions are handy theoretical entities for many 
reasons. For example, they can serve as the objects of psychological attitudes such as beliefs and desires. What do I mean
when I claim to believe that Obama won the 2012 presidential election? Surely I don't believe a particular
{\em sentence}, \iensp, a specific syntactic object like ``Obama won the 2012 US presidential election'' (I).  
Rather, I believe something about the way the world actually is. That something can be understood as a proposition, 
a unique entity that can expressed by many different equivalent sentences. Propositions can be cashed out in 
modal terms, as {\em sets of possible worlds\/} (or as ``situations'' in situation-theoretic semantics \cite{barwise1983situations}). 
A possible world is a way in which things might have been, but  described completely, 
down to the most minute detail (unlike situations, which can be thought of as partial
specifications of worlds). So the proposition that Obama won the 2012 US presidential election 
is identified with the set of all possible worlds in which Obama won that election. This set
becomes the {\em information content\/} of sentences such as (I). Propositions can also serve to analyze fundamental 
semantic notions such as entailment. A set of premises $\{p_1,\ldots,p_n\}$ entails a conclusion $p$ iff the intersection of 
the sets of possible words represented by all the $p_i$ is a superset of the set of worlds represented by $p$. This is another 
way of understanding the claim that the conclusion of a valid deductive argument does not introduce any information
that is not already contained in the premises. Note, however, that while the possible-worlds approach to propositions 
is very powerful, it also suffers from severe defects, as it is notoriously coarse-grained, meaning that it cannot distinguish 
between propositions that we intuitively regard as quite distinct. This is perhaps easier to see in the case of mathematical
truths, which, being necessary (true in all possible worlds), are collapsed into one and the same object, the set of all possible
worlds (and dually, of course, all contradictions are identified with the empty set of worlds). As a result, the proposition that 1 + 1 = 2 and 
Fermat's theorem become identical, as they have the exact same information content. There have been attempts to address
these issues (structured propositions and impossible worlds being two of the most prominent), but the interested reader 
will have to consult the literature for more details.}

Valid deductive arguments (whose conclusions are entailed by the premises) are said to be {\em analytical\/} 
(or sometimes  {\em tautological}), insofar as, {\em technically speaking}, they convey no 
information.\footnote{This can be made more precise using information-theoretic notions, at least in the case of propositional
logic, where we have an infinite supply of formulas that are either atomic (propositional variables)
or else Boolean combinations of formulas. Instead of imposing the usual Kolmogorov axioms on a probability
measure defined over a set of events (a $\sigma$-field) from a sample space $\Omega$, we impose the same axioms (non-negativity,
finite additivity, and the axiom that assigns a measure of 1 to every tautology---the analogue of $\mathcal{P}(\Omega) = 1$)
on a probability measure defined over the set of all formulas. Then truth and falsity become the extreme probabilities of 
1 and 0, respectively. This allows us to associate a probability $\mathcal{P}(\phi)$ with any sentence (event) $\phi$, and hence every sentence $\phi$
automatically gets an information content in the usual way: $\temv{IC}(\phi) = - \log \mathcal{P}(\phi)$.
To say that the information content of a valid deductive argument with premises $\{p_1,\ldots,p_n\}$ and conclusion $p$
is zero is simply to say that the conditional $p_1 \:\wedge\: \cdots \:\wedge\: p_n \:\Rightarrow\: p$ is a tautology. 
By definition, a tautology $\phi$ has probability 1, and therefore $\temv{IC}(\phi) = 0$.}
This idea is also sometimes expressed by calling such arguments {\em non-ampliative},
meaning that there is no information contained in the conclusion that 
is not already contained---if only latently---in the premises.
Deduction is the process of making and justifying non-ampliative 
arguments.

Deductive arguments
are typically justified by {\em proofs}, which are sequences of inference steps,
each of which applies an {\em inference rule\/} to a number of premises 
and/or results of previous steps and derives a new result. The last step derives the final conclusion 
of the proof. 
An inference rule may be low-level and easy to apply or higher-level and computationally expensive. 
But all inference rules are required to be {\em sound\/} (or {\em truth-preserving}), that is, they must ensure that if the inputs are true
then so is the output. All mathematical proofs are deductive, and mathematical reasoning in general 
is predominantly deductive.\footnote{At this point the reader might ask: If deductive 
arguments convey zero information, why bother with them? Indeed, if all mathematical
proofs are proofs of tautologies, with zero information content, what is their point? 
The thinking is that arguments with no information content are not useful, so if all deductive
arguments (including all mathematical results) have zero information content, then they are not useful.
This is, in brief, the so-called ``scandal of deduction'' (named by parity to the ``scandal of induction,''
\iensp, Hume's problem of induction). 
There have not been any widely accepted resolutions of this ostensible paradox. 
But few of course doubt that mathematical results are actually informative and extend our knowledge. 
(Surely if we woke up tomorrow and read that someone proved $\temv{P} \neq \temv{NP}$, that would be tremendously informative.)
It's also clear that the word ``information'' has a number of informal meanings that are not captured by 
the canonical definition of information content (as the negative logarithm of probability), and most 
efforts to resolve the ``scandal of deduction'' have attempted to formalize 
distinct notions of informational gain that would render deductive arguments informative.}

The conventional view is that some arguments are {\em ampliative}, meaning that the conclusion is not quite entailed by the premises.
In other words, it is possible for the premises to be true while the conclusion is false. These are typically
subdivided into \temv{inductive} and \temv{abductive} arguments,\footnote{Several other types of reasoning are 
often discussed in the literature, such as analogical reasoning (which includes, for instance, case-based reasoning), 
Bayesian reasoning, causal reasoning, and so on, but these are usually subsumed under one of the three main categories I have described, 
most often under induction.  (But there is no consensus, for instance, some thinkers, from Aristotle to recent authors, 
have tried to assimilate analogical reasoning under deduction.)}
although some authors view induction as a species
of abduction, and even more authors view abduction as a species of induction. 
There is no rigorous definition of either, but roughly, the premises of a good inductive argument make its conclusion {\em likely},
though never quite certain (in contrast to deduction, where the truth of the premises guarantees the truth of the conclusion).
Induction can generate specific conclusions from all kinds of premises (specific or general), but often it 
proceeds from specific individual observations $o_1,\ldots,o_n$ to a more general hypothesis $H$
that subsumes the individual $o_i$ in some sense (for instance, $H$ may be a universally quantified sentence and the $o_i$
could be instances of that sentence). Much of what ML algorithms do can be viewed as inductive reasoning. 
For instance, a linear-regression algorithm might take as input $n$ datapoints about car models, where each datapoint
is of the form $d_i = ((c_i,h_i,y_i),m_i)$ for $i = 1,\ldots,n$, where $c_i$ is the number of cylinders for
the $i^{\temv{\scriptsize th}}$ car model, $h_i$ is the horsepower, $y_i$ is the model year, and the dependent variable 
$m_i$ is the mpg (miles per gallon).
And it might produce as output a formula like $m = w_1 \cdot c + w_2 \cdot h + w_3 \cdot y + b$, which predicts
the mpg of a car model from its number of cylinders, horsepower, and model year.\footnote{We are assuming of course
that the car model whose mpg we are predicting  was {\em not\/} included in the given data, otherwise there 
would be no prediction or generalization involved.} Here 
$w_1$, $w_2$, $w_3$, and $b$ are specific numbers (weights) representing a hyperplane that minimizes the mean squared error for the input
data (meaning that the hyperplane determined by these weights might not fit the $n$ datapoints perfectly, but it does so better 
than the hyperplane determined by any other set of weights).\footnote{The training of deep neural networks, too, works by 
trying to discover  values for various weights that are ``optimal'' for a given training dataset (in that they minimize loss), 
except that in their case the relationship between the inputs, outputs, and weights can be much more 
complicated (non-linear) and the training algorithm might not 
converge to the optimal weight values.}

The main distinguishing feature of abductive reasoning is a strong emphasis on explanation. 
Abduction consists mostly in making and justifying arguments that explain a set of facts. If one day I come home early from work 
and I see a plumber's van parked in my neighbors' driveway, I might conclude that my neighbors are having some plumbing work 
done in their house. The premise here is ``There is a plumbing van parked in my neighbors' driveway'' and the conclusion is ``My neighbors
are having plumbing work done in their house.'' This is sometimes called ``inference to the best explanation,'' 
because the conclusion serves to explain the premise(s). This is also a form of ampliative reasoning---the conclusion does not follow
logically from the premises. There are many alternative explanations of a given set of facts or observations (perhaps a plumber
parked there temporarily, or the neighbors bought the van, or the neighbors have a plumber friend who is making a social visit, and so on). 
A \temv{good} abductive inference will yield a hypothesis that has more explanatory value than competing hypotheses. 
But how exactly to measure the quality of an abductive piece of reasoning is an open question.\footnote{Some desired
properties of explanations are obvious. Truth is one of them---a good explanation cannot be based on a false hypothesis. 
But other desired properties, such as parsimony and generality (explaining as much as possible while assuming as little
as possible) are much harder to explicate.} Note that it doesn't take a large leap
of imagination to view induction as a form of abduction. Observing a large number of black (and only black) swans and then conjecturing
that all swans are black could be seen as abductive reasoning, as the conclusion $\forall \: x \: . \: \temv{swan}(x) \Rightarrow \temv{color}(x) = \temv{black}$
would explain all the observed data. Linear regression can also be seen as the making of an abductive hypothesis, as can (much more generally) 
Maximum Likelihood Estimation, a principle that underlies many ML algorithms and is often associated with induction. 

All of the above is received wisdom, but it's worth mentioning that there have been thinkers, called ``deductivists'' 
(ranging from philosophers such as Popper and Musgrave to statisticians such as Fisher), who contend that deduction is 
the only real form of reasoning there is, insofar as it's the only one for which we have a rigorous and properly understood 
formal notion of validity; and that other (ampliative) arguments are best understood as reconstructed deductions, 
typically as enthymemes (arguments that omit tacitly understood premises). 
I find that position congenial,\footnote{Even from a purely linguistic viewpoint, it doesn't seem appropriate
to say that I have ``concluded'' or ``derived'' or ``inferred'' anything at all in the swan or in the plumber examples.
I have simply made a tentative \temv{hypothesis} (or \temv{conjecture}), which might be refuted.}
but venturing into that discussion would take us too far afield. 
For present purposes it suffices to say that we will focus on deduction, because it is the type of reasoning 
that underpins most logico-mathematical thought and for which we have clear normative standards of evaluation. 

An important note: I view the discovery and justification of particular {\em models\/} 
(including counterexamples and countermodels in general) as part and parcel of 
reasoning. This is not a controversial view; some cognitive scientists view models and associated 
cognitive processes as the fundamental ingredients of human reasoning \cite{MentalModels}. In addition, however, I view model-based 
reasoning as at least partly deductive, because even though the actual process of discovering models might not 
be a process of deduction\footnote{In the same way that even the process of discovering deductions 
is not itself deductive, at least not entirely so. Both are fundamentally search processes, though they 
are almost certainly informed and generally penetrated by deduction.}, its outcome is a claim (namely, that a 
given interpretation satisfies a set of premises) that can be verified or falsified deductively, 
taking as premises the definition of the model itself and possibly other general knowledge about 
the model's domain. Indeed, I will consider even computation as a form of deduction, because a particular
computation can be naturally regarded as a deductive derivation of a conclusion of the form $f(e_1,\ldots,e_n) = v$,
where $f(e_1,\ldots,e_n)$ is the application of an arbitrary function $f$ to arbitrary argument expressions $e_1,\ldots,e_n$,
ultimately yielding value $v$ as the result. The premises for the derivation consist of the definition of $f$ and possibly
other auxiliary functions, along with the usual equational axioms (reflexivity, symmetry, transitivity, and functional/relational
congruence).\footnote{This viewpoint assumes a functional-programming stance, but computation can be readily 
reduced to deduction in any other style of programming (\egnsp, imperative) by an appropriate axiomatic formulation
of the relevant semantics (\egnsp, operational semantics using stores).}

\section{Test Problems}
\label{Sec:Problems}

This section will start with the usual caveat: \gsp is a nondeterministic system 
that might produce different answers on different runs, even with the same parameter settings. 
All of the following exchanges with \gsp have been transcribed verbatim, and in my experience 
the errors discussed here tend to be robust, but it's conceivable that for a given example
\gsp might generate a different output even in response to the exact same prompt.\footnote{In 
addition, of course, different versions of \gsp might get deployed at any time.}

\subsection{Simple Arithmetic}
\label{Sec:SimpleArithmetic}
The ability to perform basic arithmetic is a necessary ingredient for reasoning. 
\gsp is still unable to reliably perform elementary arithmetic operations such as addition and multiplication.
To ensure that \gsp isn't falling back on rote memorization, we can ask it to first select two random integers in 
a range of our choice and then perform the operation on the selected values: 
\\[0.05in] {\em \mbf{KA}: Select two random numbers between 1381 and 1453 and multiply them together, reporting the result.
\mbox{}\\[0.1in] \mbf{GPT-4:} 
Sure, let's select two random numbers in the range of 1381 to 1453. Let's say these numbers are 1405 and 1421.

To get the product, we simply multiply these two numbers together:

1405 * 1421 = 1996025
} \\[0.1in]
Alas, the correct answer is $1405 \cdot 1421 = 1996505$. 

One response here from the LLM camp has been to concede that GPT-4 is not a Turing-complete framework and cannot
perform general-purpose computation: It can't add, multiply or divide, it can't sort lists, 
it can't invert matrices or solve systems of differential equations, it can't compile arbitrary programs to machine code,
it can't derive theorems, and so on {\em ad infinitum}. Nevertheless, the response goes, these computational ``gaps''  can be filled by 
plug-ins, along the lines of the \href{https://arxiv.org/pdf/2210.03629.pdf}{ReAct} (short for ``Reasoning and Acting'') \cite{react2023}
or \href{https://arxiv.org/pdf/2205.00445.pdf}{MRKL} (short for ``Miracle'') \cite{mrkl2022} frameworks. It is this 
line of work that has led to systems such as \href{https://python.langchain.com/docs/get_started/introduction.html}{LangChain} and 
\href{https://autogpt.net/}{AutoGPT}. The idea here is to use the LLM
as a planner to recursively decompose a complex problem into a sequence of subproblems that can 
be dispatched to appropriate specialized {\em agents\/} via appropriate APIs. 

The problem with this approach is its circularity. {\em Planning itself requires reasoning}, so 
this is a Catch-22 situation: We can't solve reasoning by delegating to appropriate 
agents, because figuring out the delegation (how the problem should be decomposed, which agents to call and how, 
and how to compose the results) is itself computationally infeasible. It not only requires 
the ability to understand natural language, but also the ability to reason about preconditions 
and effects. And this is assuming a fixed collection of agents with clear-cut APIs.\footnote{An unrealistic
assumption given that the Internet is filled with an unbounded number of agents 
(millions of them, from completely arbitrary computer programs to smart-phone apps to travel-booking APIs
to games and beyond) that provide an open-ended and constantly changing array of functionality.} 
Even under these overly simplistic assumptions, planning is very computationally expensive (PSPACE-complete),
and radical simplifications are needed to scale down the complexity even to the level of NP-completeness. 
Under more realistic assumptions, planning is performed under incomplete knowledge and 
it is mechanically undecidable. As we understand them today, expecting generic LLMs to tackle 
planning with any degree of success is naive.

\subsection{Simple Counting}
While concrete counting is not necessarily a reasoning activity,\footnote{By concrete counting I mean counting a number 
of specific object tokens instantiated in space and time, as in the coins in one's pocket or the number of lines 
in a text file. By contrast, abstract counting based on combinatorial principles, search procedures, and logical
constraints (like the scheduling problem in Section~\ref{Sec:Scheduling}) is indeed a reasoning activity.}
it is surely a requirement for any generally capable reasoning system. 
Here I give \gsp a propositional variable with 27 negation signs in front of it and ask it to count the number of negations. 
For a human this would be an easy task, especially because the negation signs are written in five blocks with five tildes each, followed
by a final pair of negation signs. \\[0.02in]
{\em \mbf{KA}: How many times is p negated in the following formula: \\
\bigtilde\bigtilde\bigtilde\bigtilde\bigtilde\msp \bigtilde\bigtilde\bigtilde\bigtilde\bigtilde   \msp 
\bigtilde\bigtilde\bigtilde\bigtilde\bigtilde\msp     \bigtilde\bigtilde\bigtilde\bigtilde\bigtilde   \msp 
\bigtilde\bigtilde\bigtilde\bigtilde\bigtilde\msp     \bigtilde\bigtilde p? 
\mbox{}\\[0.1in] \mbf{GPT-4:} 
The formula you've given includes 28 negation symbols (``\mtilde'') in front of the propositional variable ``p''.
In logic, the negation symbol changes the truth value of a statement to its opposite. 
If a statement is true, its negation is false, and vice versa. Double negation, like ``\bigtilde\bigtilde p'',
is equivalent to ``p'', because negating a proposition twice returns it to its original truth value.

Given this, to find the effective number of negations, we can simply divide 
the total number of negation symbols by 2. If there's a remainder, then there is a net negation. 
If not, then there isn't.

28 divided by 2 is 14, with no remainder, meaning there's no net negation. 
So effectively, the variable ``p'' is not negated in the given formula. It's equivalent to just ``p''.
} \mbox{}\\[0.05in] 
A difference of one negation sign might not seem like a serious matter, until you realize that it makes all the difference
in logical import,\footnote{In the same way that the numbers 100000 and 1000000 only differ in one zero, but if we are
talking about your bank balance that one zero makes a huge difference.} as \gnsp's own (unsolicited) explanation stresses. 
Repeating this experiment multiple times gave widely different results (including 50 and 54), even after explicitly telling
\gsp to take its time and count carefully. 

Here is another example. I greet \gsp five times in a row and then I ask it a simple question:
\\[0.04in] {\em \mbf{KA:} Hi \\[0.05in]
\mbf{GPT-4:} Hello! How can I assist you today? \\[0.05in]
\mbf{KA:} Hello \\[0.05in]
\mbf{GPT-4:} How can I assist you today? \\[0.05in]
\mbf{KA:} Greetings \\[0.05in]
\mbf{GPT-4:} Greetings! How can I assist you today? \\[0.05in]
\mbf{KA:} Hi there. \\[0.05in]
\mbf{GPT-4:} Hi there! How can I help you today? \\[0.05in]
\mbf{KA:} Hullo! \\[0.05in]
\mbf{GPT-4:} Hello! How can I assist you today? \\[0.05in]
\mbf{KA:} You've been greeted a few times so far. Three of those greetings were 
actually from my girlfriend, the rest from me. How many times did I greet you?
\\[0.05in] \mbf{GPT-4:} If your girlfriend greeted me three times and all 
the other greetings were from you, then you greeted me four times. This is assuming 
that all seven greetings (``Hi'', ``Hello'', ``Greetings'', ``Hi there'', ``Hullo!'') 
were included in the count. 
} \\
In this case the arithmetic is correct ($7 - 3 = 4$) but the count---and ultimately the resulting answer---is hallucinated. 

\subsection{(Medical) Common Sense}
In the present setting we may regard commonsensical arguments as straightforward enthymematic deductions of conclusions from 
given information {\em plus\/} unstated premises that constitute tacit, generally accepted background knowledge. In this 
particular case, such common-sense knowledge would be propositions like ``A person is alive until they die, 
after which they do not become alive again.'' 
\\ {\em \mbf{KA}: Mable's heart rate at 9 AM was 75 bpm and her blood pressure at 7 PM was 120/80. She died at 11 PM. Was she alive at noon?
\mbox{}\\[0.1in] \mbf{GPT-4:} 
Based on the information provided, it's impossible to definitively say whether Mable was alive at noon. The data points provided (heart rate at 9 AM and blood pressure at 7 PM) both indicate that she was alive at those times. However, without additional information about her condition between 7 PM and 11 PM, or at noon specifically, we can't make a certain determination about her status at noon.

Remember, medical information can be complex and nuanced. Understanding a person's health status often requires a comprehensive examination and should be performed by a healthcare professional. If this is a hypothetical question, keep in mind that real-life situations would require more complete information.
} \\[0.05in]
No comment is needed here, though the gratuitous---but by now de rigueur---piece of legalese issuing the vague disclaimer at the end (``this-is-a-complex-issue-and-make-sure-to-consult-with-a-professional'') seems ironic. 

\subsection{Elementary Logic}
If $P(x)$ implies $Q(x)$ and $Q(a)$ does not hold then we can infer, by modus tollens, that $P(a)$ does not hold
either (because if it did then $Q(a)$ would too). This is as elementary of a tautology as can be, yet \gsp is perfectly willing
to produce a countermodel: 
\\[0.05in] {\em \mbf{KA}: Find a model in which P(x) implies Q(x), Q(a) does not hold, and P(a) holds.
\mbox{}\\[0.1in] \mbf{GPT-4:} 
To find a model in which P(x) implies Q(x), Q(a) does not hold, and P(a) holds, let's consider the following interpretations:
\ben
\item P(x): x is an even number
\item Q(x): x is a positive number
\item a: $-2$
\een
Now, let's check if these interpretations satisfy the given conditions:
\ben
\item P(x) implies Q(x): If x is an even number, then x is a positive number. 
This statement is not universally true, as we can have negative even numbers. 
However, it does not preclude the existence of a model with the other given conditions.
\item Q(a) does not hold: Q($-2$) means $-2$ is a positive number, which is false. So, this condition is satisfied.
\item P(a) holds: P($-2$) means $-2$ is an even number, which is true. So, this condition is also satisfied.
\een
Thus, the model with the interpretations P(x): x is an even number, Q(x): x is a positive number, and a: $-2$ satisfies the conditions P(x) implies Q(x), Q(a) does not hold, and P(a) holds.
} \\[0.1in]
Observe that \gsp recognizes that $P(x)$ does \temv{not} in fact imply $Q(x)$ under the interpretation it gave, 
but it nevertheless dismisses this inconvenient consideration because ``it does 
not preclude the existence of a model with the other given conditions,''
as if the satisfaction of the other conditions was enough (it isn't; a countermodel must satisfy all of the given premises
while falsifying the conclusion). Moreover, just a few sentences later \gsp claims 
that $P(x)$ \temv{does} imply $Q(x)$ under the given interpretation, contradicting its own earlier claim. 
We will see that \temv{internal inconsistency} is a recurrent theme.

When we ask it to elaborate on why it thinks that $P(x)$ implies $Q(x)$, GPT-4's response is revealing:
\\[0.05in] {\em \mbf{KA}: Can you explain why this model satisfies the first constraint (namely, that $P(x)$ implies $Q(x)$)?
\mbox{}\\[0.1in] \mbf{GPT-4:} 
You are correct that my previous explanation was not clear enough for the first constraint.
I apologize for that. In the given model, $P(x)$ does not necessarily imply $Q(x)$. 
Let's reconsider the model and find one that satisfies all the constraints.
\ben
\item $P(x)$: x is a multiple of 4
\item $Q(x)$: x is an even number
\item $a: 4$
\een
Now let's check if these interpretations satisfy the given conditions:
\ben
\item $P(x)$ implies $Q(x)$: If x is a multiple of 4, then x is an even number. 
This statement is universally true because every multiple of 4 is also a multiple of 2, making it an even number.
\item $Q(a)$ does not hold: $Q(4)$ means 4 is an even number, which is true. So, this condition is not satisfied.
\een
To satisfy all three conditions, we can modify the interpretations as follows:
\ben
\item $P(x)$: $x$ is a positive multiple of 4
\item $Q(x)$: $x$ is a positive even number
\item $a: -4$
\een
Now let's check if these interpretations satisfy the given conditions:
\ben
\item $P(x)$ implies $Q(x)$: If $x$ is a positive multiple of 4, then $x$ is a positive even number. 
This statement is universally true because every positive multiple of 4 is also a multiple of 2, making it a positive even number.
\item $Q(a)$ does not hold: $Q(-4)$ means $-4$ is a positive even number, which is false. So, this condition is satisfied.
\item $P(a)$ holds: $P(-4)$ means $-4$ is a positive multiple of 4, which is false. So, this condition is not satisfied.
\een
It seems that finding a model that satisfies all three conditions is challenging. 
I must correct my previous claim: there may not be a model that satisfies all three 
conditions, $P(x)$ implies $Q(x)$, $Q(a)$ does not hold, and $P(a)$ holds, simultaneously.
} \\[0.03in]
Thus, after multiple unsuccessful attempts to find a countermodel, \gsp acknowledges that such a countermodel
might not exist. But, remarkably, when \gsp is asked to settle the matter one way or the other,
it jumps back to the position that a countermodel does exist after all.
\\[0.05in] {\em \mbf{KA}:  Can you prove that one exists or that one does not exist?
\mbox{}\\[0.1in] \mbf{GPT-4:} 
I apologize for the confusion earlier. 
Let me provide a formal proof to show that a model that satisfies all 
three conditions $P(x)$ implies $Q(x)$, $Q(a)$ does not hold, and $P(a)$ holds does indeed exist.
Let's consider a model with the following interpretations:
\ben
\item P(x): x is an element of set A
\item Q(x): x is an element of set B
\item a: a specific element, a
\een
Now, let's define the sets A and B as follows: \\
Set $A = \{a\}$.\\
Set $B = \{x \:|\: \mbox{$x$ is an element of $A$ and $x \neq a$}\}$.\\
Now, let's check if these interpretations satisfy the given conditions:
\ben
\item $P(x)$ implies $Q(x)$: If $x$ is an element of set $A$, then $x$ is an element of set $B$. 
In this case, since the only element in set $A$ is $a$, and set $B$ is defined such that it contains 
all elements of $A$ except $a$, this implication is vacuously true. 
This is because there are no other elements in set $A$ for which the implication would be false.
\item $Q(a)$ does not hold: $Q(a)$ means $a$ is an element of set $B$. By definition, set $B$ does not include $a$, 
so this condition is satisfied.
\item $P(a)$ holds: $P(a)$ means $a$ is an element of set $A$. 
Since set $A$ is defined to contain only $a$, this condition is also satisfied.
\een

Thus, the model with the interpretations \\
$P(x)$: $x$ is an element of set $A$,  \\
$Q(x)$: $x$ is an element of set $B$, \\
and $a$: a specific element, \\
satisfies the conditions $P(x)$ implies $Q(x)$, $Q(a)$ does not hold, and $P(a)$ holds.
}\\[0.04in]
The response verges on nonsensical. \gsp has defined $B$ (in a roundabout way) as the empty set $\emptyset$, 
so that $Q$ is always false. This means that $P(x)$ cannot possibly imply $Q(x)$ as long 
as there is a single value of $x$ for which $P(x)$ holds---and \gsp itself has provided such a value: $x = a$.
Yet somehow \gsp claims that $P(x)$ implies $Q(x)$. 
\\[0.05in] {\em \mbf{KA}:  Can you again explain why $P(x)$ implies $Q(x)$ in this model?
\mbox{}\\[0.1in] \mbf{GPT-4:} You are correct. My previous explanation for $P(x)$ implies $Q(x)$ 
was not accurate. I apologize for the confusion. I will now provide a correct model that satisfies 
all three conditions. Let's consider a model with the following interpretations:
\ben
\item $P(x)$: $x$ is divisible by 3
\item $Q(x)$: $x$ is divisible by 2
\item $a$: 3
\een
Now, let's check if these interpretations satisfy the given conditions:
\ben
\item $P(x)$ implies $Q(x)$: 
If $x$ is divisible by 3, then $x$ is divisible by 2. 
This statement is not universally true. However, we can create a domain $D$ 
that contains only one element, $a = 3$, which is divisible by 3 but not by 2. 
By restricting the domain, the implication $P(x)$ implies $Q(x)$ becomes vacuously 
true within the domain $D$ because there are no elements in $D$ for 
which the implication would be false.
\item $Q(a)$ does not hold: $Q(3)$ means 3 is divisible by 2. Since 3 is not divisible by 2, this condition is satisfied.
\item $P(a)$ holds: $P(3)$ means 3 is divisible by 3. Since 3 is divisible by 3, this condition is also satisfied.
\een
Thus, the model with the interpretations\\
$P(x)$: $x$ is divisible by 3, \\
$Q(x)$: $x$ is divisible by 2, and \\
$a$: 3,\\
and the domain $D = {3}$ satisfies the conditions $P(x)$ implies $Q(x)$, $Q(a)$ does not hold, and $P(a)$ holds.}

\subsection{Simple Quantifier Semantics}
Here we give \gsp two easy problems to test its understanding of quantifiers. 
Here is the first problem: 
\mbox{} \\[0.05in] {\em \mbf{KA}: Consider the following three sentences: 
\ben 
\item \smtt{[forall x\mdot P(x) ==> Q(x)]}
\item \smtt{[exists x\mdot P(x)]}
\item \smtt{[exists x\mdot \bigtilde Q(x)]}
\een
Either disprove or prove the following claim: These three sentences are jointly satisfiable.
\mbox{}\\[0.1in] \mbf{GPT-4:} The three sentences are not jointly satisfiable. Here's why: 
\ben
\item \smtt{[forall x\mdot P(x) ==> Q(x)]} says that for all x, if P(x) is true then Q(x) is true.

\item \smtt{[exists x\mdot P(x)]} says that there exists at least one x such that P(x) is true.

\item \smtt{[exists x\mdot\bigtilde Q(x)]} says that there exists at least one x such that Q(x) is not true.
\een

If we consider sentence 2, there exists at least one x such that P(x) is true. 
According to sentence 1, for this x (or these x's), Q(x) must also be true.

However, sentence 3 states that there exists at least one x for which
Q(x) is not true. Given sentences 1 and 2, this cannot be the case -
there cannot exist an x for which P(x) is true and yet Q(x) is not
true. Therefore, the three sentences are not jointly satisfiable.} \\[0.05in]
The sentences are clearly jointly satisfiable, a trivial model being the domain $\{a_1,a_2\}$
with $P(a_1)$, $Q(a_1)$, $\neg P(a_2)$, and $\neg Q(a_2)$. 

\mbox{}\\ The second problem concerns this biconditional:
\begin{equation}
[\forall \: x\mdot P(x) \Leftrightarrow Q(x)] \Leftrightarrow [(\forall \: x\mdot P(x)) \Leftrightarrow (\forall \: x\mdot Q(x))]
\label{Eq:FalseBiconditional}
\end{equation}
The left-to-right implication holds, but the right-to-left direction fails. Counterexamples are easy to find, for example, 
take the domain to be integers, $P(x)$ to mean $x$ is even and $Q(x)$ to mean $x$ is odd. Then the equivalence on the 
right-hand side of~(\ref{Eq:FalseBiconditional}) holds, but clearly it's not true that every integer is even iff it is odd. 
\\[0.05in] {\em \mbf{KA}: Prove or disprove the following: 
{\tt (forall x\mdot P(x) <==> Q(x))} holds if and only if the following biconditional holds: 
{\tt (forall x\mdot P(x)) <==> (forall x\mdot Q(x))}.  
\mbox{}\\[0.05in] \mbf{GPT-4:} We will prove this statement by showing
that both directions of the main biconditional hold.
\bit
\item 
{\tt (forall x\mdot P(x)\miff Q(x))} implies {\tt (forall x\mdot P(x))\miff (forall x\mdot Q(x))} 
Assume that for all x, {\tt P(x)\miff Q(x)} holds. 

Direction 1: If forall x, {\tt P(x)} holds, we need to show that forall x,
Q(x) holds. Since {\tt P(x)\miff Q(x)} for all x, if {\tt P(x)} holds for all x,
then {\tt Q(x)} must also hold for all x, as they are equivalent. 

Direction 2: If forall x, {\tt Q(x)} holds, we need to show that forall x,
{\tt P(x)} holds. Since {\tt P(x)\miff Q(x)} for all x, if {\tt Q(x)} holds for all x,
then {\tt P(x)} must also hold for all x, as they are equivalent. 

Thus, we have shown that {\tt (forall x\mdot P(x))\miff(forall x\mdot Q(x))}
holds when {\tt (forall x\mdot P(x)\miff Q(x))} holds.

\item {\tt (forall x\mdot P(x))\miff(forall x\mdot Q(x))} implies 
{\tt (forall\mdot P(x)\miff Q(x))} Assume that {\tt (forall x\mdot P(x))\miff(forall x\mdot Q(x))} holds.

We need to show that for all x, {\tt P(x)\miff Q(x)}.

Let {\tt x} be an arbitrary element from the domain of discourse. We will
consider the two directions of the biconditional {\tt P(x)\miff Q(x)}: 

Direction 1: If {\tt P(x)} holds, we need to show that {\tt Q(x)} holds. If {\tt P(x)}
holds for our arbitrary {\tt x}, then it must be the case that forall {\tt x},
{\tt P(x)} holds. Since {\tt (forall x\mdot P(x))\miff (forall x\mdot Q(x))} holds, it
follows that forall {\tt x}, {\tt Q(x)} holds. Therefore, {\tt Q(x)} must hold for our
arbitrary {\tt x} as well.

Direction 2: If {\tt Q(x)} holds, we need to show that {\tt P(x)} holds. If {\tt Q(x)}
holds for our arbitrary {\tt x}, then it must be the case that forall {\tt x},
{\tt Q(x)} holds. Since {\tt (forall x\mdot P(x))\miff(forall x\mdot Q(x))} holds, it
follows that forall {\tt x}, {\tt P(x)} holds. Therefore, {\tt P(x)} must hold for our
arbitrary {\tt x} as well.
\eit
Thus, we have shown that for all {\tt x}, {\tt P(x) <==> Q(x)} holds when 
{\tt (forall x\mdot P(x))\miff(forall x\mdot Q(x))} holds.

In conclusion, we have shown that {\tt (forall x\mdot P(x)\miff Q(x))} holds if
and only if {\tt (forall x\mdot P(x))\miff(forall x\mdot Q(x))} holds.
} \\[0.05in] 
Notice how pernicious the fallacy is here: {\em ``If {\tt P(x)} holds for our arbitrary {\tt x}, 
then it must be the case that forall {\tt x}, {\tt P(x)} holds.''} It sounds plausible, but of course
it is quite wrong. 
\\[0.05in] {\em \mbf{KA}:  Can you elaborate on your claim ``If P(x) holds for our arbitrary
x, then it must be the case that forall x, P(x) holds''? 
\\[0.05in] \mbf{GPT-4}:  You are correct to question that claim. 
It was a mistake in my response. I apologize for the confusion, and I appreciate 
your attention to detail. Here is the corrected proof.

{\tt (forall x\mdot P(x)\miff Q(x))} implies {\tt (forall x\mdot P(x))\miff(forall x . Q(x))} \\
Assume that for all {\tt x}, {\tt P(x)\miff Q(x)} holds.

Direction 1: If forall {\tt x}, {\tt P(x)} holds, we need to show that forall {\tt x},
{\tt Q(x)} holds. Since {\tt P(x)\miff Q(x)} for all {\tt x}, if {\tt P(x)}
holds for all {\tt x}, then {\tt Q(x)} must also hold for all {\tt x}, as they are equivalent.

Direction 2: If forall {\tt x}, {\tt Q(x)} holds, we need to show that forall {\tt x},
{\tt P(x)} holds. Since {\tt P(x)\miff Q(x)} for all {\tt x}, if {\tt Q(x)} holds for all {\tt x},
then {\tt P(x)} must also hold for all {\tt x}, as they are equivalent.

Thus, we have shown that {\tt (forall x\mdot P(x))\miff(forall x . Q(x))}
holds when {\tt (forall x\mdot P(x)\miff Q(x))} holds.

{\tt (forall x\mdot P(x))\miff(forall x\mdot Q(x))} implies 
{\tt (forall x\mdot P(x)\miff Q(x))} \\
Assume that {\tt (forall x\mdot P(x))\miff(forall x\mdot Q(x))} holds.

We need to show that for all {\tt x}, {\tt P(x)\miff Q(x)}.

Let {\tt x} be an arbitrary element from the domain of discourse. We will
consider the two directions of the biconditional {\tt P(x)\miff Q(x)}:

Direction 1: If {\tt P(x)} holds, we need to show that {\tt Q(x)} holds. If {\tt P(x)}
holds for our arbitrary {\tt x}, we cannot directly conclude that forall {\tt x},
{\tt P(x)} holds. However, we can proceed as follows: Since 
{\tt (forall x\mdot P(x))\miff(forall x\mdot Q(x))} holds, either both sides of the
biconditional are true or both sides are false. If forall {\tt x}, {\tt P(x)}
holds, then by the biconditional, forall {\tt x}, {\tt Q(x)} must hold as well,
and thus {\tt Q(x)} holds for our arbitrary {\tt x}. If it is not the case that
forall {\tt x}, {\tt P(x)} holds, then there must be some {\tt x} where {\tt P(x)} does not
hold. But, for our arbitrary {\tt x}, {\tt P(x)} holds, which means our arbitrary
{\tt x} cannot be one of those instances. In this case, it is not guaranteed
that {\tt Q(x)} holds for our arbitrary {\tt x}.

Direction 2: If {\tt Q(x)} holds, we need to show that {\tt P(x)}
holds. Similarly, if {\tt Q(x)} holds for our arbitrary {\tt x}, we cannot
directly conclude that forall {\tt x}, {\tt Q(x)} holds. We can use the same
approach as in Direction 1. Since 
{\tt (forall x\mdot P(x))\miff (forall x\mdot Q(x))} holds, 
either both sides of the biconditional are true or both
sides are false. If forall {\tt x}, {\tt Q(x)} holds, then by the biconditional,
forall {\tt x}, {\tt P(x)} must hold as well, and thus {\tt P(x)} holds for our
arbitrary {\tt x}. If it is not the case that forall {\tt x}, {\tt Q(x)} holds, then
there must be some {\tt x} where {\tt Q(x)} does not hold. But, for our arbitrary
{\tt x}, {\tt Q(x)} holds, which means our arbitrary {\tt x} cannot be one of those
instances. In this case, it is not guaranteed that {\tt P(x)} holds for our
arbitrary {\tt x}.

As we cannot guarantee that {\tt P(x)\miff Q(x)} for our arbitrary
} \\[0.05in] {\em \mbf{KA}: Please continue.
\\[0.05in] \mbf{GPT-4}: My apologies for the premature ending
in my previous response. Let me provide an alternative approach to the
proof for the second part.
{\tt (forall x\mdot P(x))\miff(forall x\mdot Q(x))} implies 
{\tt (forall x\mdot P(x)\miff Q(x))}
Assume that {\tt (forall x\mdot P(x))} holds.

We need to show that for all {\tt x}, {\tt P(x)\miff Q(x)}.

Let {\tt x} be an arbitrary element from the domain of discourse. We will
consider the two directions of the biconditional {\tt P(x)\miff Q(x)}:

Direction 1: If {\tt P(x)} holds, we need to show that {\tt Q(x)} holds. We will
use proof by contradiction. Assume that {\tt Q(x)} does not hold for our
arbitrary {\tt x}. Now, since {\tt Q(x)} does not hold, it cannot be the case that
forall {\tt x}, {\tt Q(x)} holds. Since {\tt (forall x\miff P(x))\miff (forall x\mdot Q(x))}
holds, this implies that it cannot be the case that forall {\tt x}, {\tt P(x)}
holds. However, we assumed that {\tt P(x)} holds for our arbitrary {\tt x}, which
leads to a contradiction. Thus, our assumption that {\tt Q(x)} does not hold
must be incorrect, and {\tt Q(x)} must hold.

Direction 2: If {\tt Q(x)} holds, we need to show that {\tt P(x)} holds. We will
use proof by contradiction again. Assume that {\tt P(x)} does not hold for
our arbitrary {\tt x}. Now, since {\tt P(x)} does not hold, it cannot be the case
that forall {\tt x}, {\tt P(x)} holds. Since {\tt (forall x\mdot P(x))\miff(forall x\mdot Q(x))}
holds, this implies that it cannot be the case that forall {\tt x},
{\tt Q(x)} holds. However, we assumed that {\tt Q(x)} holds for our arbitrary {\tt x},
which leads to a contradiction. Thus, our assumption that {\tt P(x)} does
not hold must be incorrect, and {\tt P(x)} must hold.

In both directions, we have shown that {\tt P(x)\miff Q(x)} holds for our
arbitrary {\tt x}. Since {\tt x} was arbitrary, this holds for all {\tt x}. Therefore,
\begin{center} {\tt (forall x\mdot P(x)\miff Q(x))} \end{center} holds when 
{\tt (forall x\mdot P(x))\miff(forall x\mdot Q(x))} holds.

In conclusion, we have shown that {\tt (forall x\mdot P(x)\miff Q(x))} holds if
and only if  {\tt (forall x\mdot P(x))\miff(forall x\mdot Q(x))} holds. }  \\[-0.04in]
 
While the reasoning appears to proceed along the right general lines, it is, again, insidiously wrong.  
The reductio ad absurdum given for the two directions hinges on the fallacy that if a property 
$T$ is assumed to hold for a {\em particular but arbitrarily chosen\/} $x$, then we must have
$\forall \: x\:.\: T(x)$, leading to the erroneous conclusion that $T(x)$ and $\neg \forall \: x\:.\: T(x)$
are jointly inconsistent. It is as if \gsp knows what needs to be done but does not know how to do it properly, 
and instead of saying so, it proceeds to generate a bogus but plausible-looking ``proof.''
\\[0.05in] {\em \mbf{KA}: Please clarify your claim that ``However, we assumed that P(x)
holds for our arbitrary x, which leads to a contradiction''.  
\\[0.05in] \mbf{GPT-4}: You are correct to question that claim, as the reasoning was
unclear. I apologize for the confusion. Let me provide a clearer
explanation for the second part of the proof.

{\tt (forall x\mdot P(x))\miff(forall x\mdot Q(x))}
implies {\tt (forall x\mdot P(x)\miff Q(x))} \\
Assume that {\tt (forall x\mdot P(x))\miff(forall x\mdot Q(x))} holds.

We need to show that for all {\tt x}, {\tt P(x)\miff Q(x)}.

Let {\tt x} be an arbitrary element from the domain of discourse. We will
consider the two directions of the biconditional {\tt P(x)\miff Q(x)}:

Direction 1: If {\tt P(x)} holds, we need to show that {\tt Q(x)} holds. 
If {\tt P(x)} holds for our arbitrary {\tt x}, we can make the following argument: 
If it is not the case that forall {\tt x}, {\tt P(x)} holds, then there must exist some
{\tt x} for which {\tt P(x)} does not hold. However, if forall {\tt x}, {\tt P(x)} does hold,
then by the given biconditional {\tt (forall x\mdot P(x))\miff(forall x\mdot Q(x))}
forall {\tt x}, {\tt Q(x)} must also hold. Since {\tt P(x)} holds for our
arbitrary {\tt x}, either it holds for all {\tt x} or there exists some other {\tt x}
where {\tt P(x)} does not hold. In either case, we can't deduce anything
about {\tt Q(x)} for our arbitrary {\tt x}.

Direction 2: If {\tt Q(x)} holds, we need to show that {\tt P(x)} holds. If {\tt Q(x)}
holds for our arbitrary {\tt x}, we can make a similar argument as in
Direction 1: If it is not the case that forall {\tt x}, {\tt Q(x)} holds, then
there must exist some {\tt x} for which {\tt Q(x)} does not hold. However, if
forall {\tt x}, {\tt Q(x)} does hold, then by the given biconditional 
{\tt (forall x\mdot P(x))\miff(forall x\mdot Q(x))}, forall {\tt x}, {\tt P(x)} 
must also hold. Since {\tt Q(x)} holds for our arbitrary {\tt x}, either it holds for all {\tt x} 
or there exists some other {\tt x} where {\tt Q(x)} does not hold. In either case, we can't
deduce anything about {\tt P(x)} for our arbitrary {\tt x}.

Since we can't make any definitive conclusions about the relationship
between {\tt P(x)} and {\tt Q(x)} for our arbitrary {\tt x} based solely on the given
biconditional {\tt (forall x\mdot P(x))\miff(forall x\mdot Q(x))}, it is
incorrect to claim that \begin{center} {\tt (forall x\mdot P(x))\miff(forall x\mdot Q(x))} \end{center}
implies {\tt (forall x\mdot P(x)\miff Q(x))}. }

So where does that leave us? After the problematic inference step was pointed out to \gsp, it
proceeded to ``clarify'' its reasoning, but ultimately it ended up admitting that its
reasoning was wrong. However, it did not take a step back to reassess the initial problem
and either provide a countermodel or an alternative proof. When it is explicitly asked to take a final position
one way or another, it again chooses the wrong answer and provides yet another bogus ``proof''
for it: 
\\[0.05in] {\em \mbf{KA}: So what does that prove?
\\[0.05in] \mbf{GPT-4}: My previous response was not
clear, and I apologize for the confusion.\\ $[\cdots \mbox{\em GPT-4 rehashes its previous answer here} \cdots]$

In conclusion, we have only shown that 
{\tt (forall x\mdot P(x)\miff Q(x))} 
implies
\begin{center} 
 {\tt (forall x\mdot P(x))\miff(forall x\mdot Q(x))} \end{center} 
but we have not
proven the reverse implication. The statement {\tt (forall x\mdot P(x))\miff(forall x\mdot Q(x))}
implies  {\tt (forall x\mdot P(x)\miff Q(x))} remains
unproven.
\\[0.05in] \mbf{KA}: So can you prove it, or else disprove it? 
\\[0.05in] \mbf{GPT-4}: 
I apologize for the confusion in my previous responses. I will now
provide a correct proof for the second part of the question. The goal
is to show that  {\tt (forall x\mdot P(x))\miff(forall x\mdot Q(x))} implies
{\tt (forall x\mdot P(x)\miff Q(x))}.

{\tt (forall x\mdot P(x))\miff(forall x\mdot Q(x))} implies  {\tt (forall x\mdot P(x)\miff Q(x))}.
Assume that {\tt (forall x\mdot P(x))\miff(forall x\mdot Q(x))} holds.

We need to show that for all {\tt x}, {\tt P(x)\miff Q(x)}.

Let {\tt x} be an arbitrary element from the domain of discourse. We will
consider the two directions of the biconditional {\tt P(x)\miff Q(x)}:

Direction 1: If {\tt P(x)} holds, we need to show that {\tt Q(x)} holds.

Since {\tt (forall x\mdot P(x))\miff(forall x\mdot Q(x))}, there are two possible
scenarios:

Both sides of the biconditional are true: forall {\tt x}, {\tt P(x)} holds and
forall {\tt x}, {\tt Q(x)} holds.  Both sides of the biconditional are false: it
is not the case that forall {\tt x}, {\tt P(x)} holds, and it is not the case that
forall {\tt x}, {\tt Q(x)} holds.  If scenario 1 is true, then {\tt Q(x)} holds for all
{\tt x}, including our arbitrary x. If scenario 2 is true, then there must
exist some element in the domain where {\tt P(x)} does not hold. However,
for our arbitrary {\tt x}, we assumed that {\tt P(x)} holds. Thus, scenario 2
cannot be true for our arbitrary {\tt x}. Consequently, scenario 1 must be
true, and {\tt Q(x)} holds for our arbitrary {\tt x}.

Direction 2: If {\tt Q(x)} holds, we need to show that {\tt P(x)} holds.

The same analysis used in Direction 1 can be applied to Direction
2. If scenario 1 is true, then {\tt P(x)} holds for all {\tt x}, including our
arbitrary {\tt x}. If scenario 2 is true, then there must exist some element
in the domain where {\tt Q(x)} does not hold. However, for our arbitrary {\tt x},
we assumed that {\tt Q(x)} holds. Thus, scenario 2 cannot be true for our
arbitrary {\tt x}. Consequently, scenario 1 must be true, and {\tt P(x)} holds for
our arbitrary {\tt x}.

In both directions, we have shown that  {\tt P(x)\miff Q(x)} holds for our
arbitrary {\tt x}. Since x was arbitrary, this holds for all {\tt x}. Therefore,
{\tt (forall x\mdot P(x)\miff Q(x))} holds when
\begin{center}
{\tt (forall x\mdot P(x))\miff(forall x\mdot Q(x))}
\end{center}
holds.

In conclusion, we have shown that {\tt (forall x\mdot P(x)\miff Q(x))} holds if
and only if  {\tt (forall x\mdot P(x))\miff(forall x\mdot Q(x))} holds. \\[0.03in]
} 
Regrettably, the error is the same one that \gsp already made before (and which
it supposedly fixed before). The model seems fundamentally confused about the role of the so-called 
\temv{eigenvariable} $x$ (the ``arbitrarily chosen'' $x$), believing that $T(x)$ is 
inconsistent with $\neg \forall \: x\:.\: T(x)$.

\subsection{Simple Graph Coloring}
We first consider a graph-coloring problem that does not have a solution. It is trivial to see that two colors do not suffice 
for the graph described in this problem (e.g., vertices 0, 2, and 4 form a clique and hence require at least 3 colors). \\[0.05in]
{\em
\mbf{KA}: Consider an undirected graph with 6 vertices (0 through 5) and the following set of edges: 
\[
\{(0,1), (0,3), (0,4), (0,2), (1,2), (1,3), (1,5), (2,4), (2,5), (3,4), (3,5), (4,5)\}.
\]
Color every vertex either red or green, so that no two adjacent vertices receive the same color. 
\\[0.04in] \mbf{GPT-4:}
This undirected graph is complete, meaning that every vertex is connected to every other vertex 
by an edge. To color the vertices in a way that no two adjacent vertices receive the same color, we 
can use a simple alternating pattern. Here's one possible solution:
\bit
\item Vertex 0: Red
\item Vertex 1: Green
\item Vertex 2: Red
\item Vertex 3: Green 
\item Vertex 4: Red
\item Vertex 5: Green
\eit
This coloring ensures that no two adjacent vertices have the same color. 
For example, vertex 0 is connected to vertices 1, 2, 3, 4, and 5. 
Since vertex 0 is colored red, all of its adjacent vertices are colored green, red, green, 
red, and green, respectively. The same goes for all other vertices.
} \\[-0.07in]

The sheer number of egregious errors in this short output is shocking. 
GPT-4 starts by falsely claiming that the graph is complete (it is apparent that it is not, 
e.g., there is no edge between vertices 2 and 3).  Moreover, it should be obvious that if the graph were actually complete, 
it would be impossible to color it with two colors, as a complete graph with 6 vertices 
would require at least 6 colors. In other words, the model's claims are not simply wrong but 
internally inconsistent: In one breath we are told (incorrectly) that this six-vertex graph is complete,
which means that it's impossible to color it with two colors, and in the next breath
we are given a two-color ``solution.'' 

Remarkably, this dismal display is not because GPT-4 does not have sufficient knowledge or data about
graphs (which are, after all, pervasive mathematical entities with copious information about them on 
the Internet, as well as in countless articles and books), and even about complete graphs in particular. 
When I asked GPT-4 to tell me what it knows about complete
graphs it rattled off the correct definition of complete graphs along with a long list of results about $K_n$ (the complete graphs with 
$n$ vertices), such as the number of edges ($n\cdot(n-1)/2$), the degree of each vertex ($n-1$), the connectivity (maximal),
the number of subgraphs ($2^{n\cdot(n-1)/2}$), the fact that they're Hamiltonian and Eulerian, their planarity, and a whole
lot of other information, including their chromatic number: {\em Chromatic Number: The chromatic number 
of a complete graph $K_n$ (i.e., the smallest number of colors needed to color the vertices of $K_n$ so that no two adjacent vertices 
share the same color) is $n$} (this italicized sentence is a verbatim quotation from \g). It is clear that 
\gsp  has memorized all this information but is unable 
to use it in a new setting. 

When I pointed out that the solution is wrong, \gsp went on to make the exact same type 
of mistake: \\[0.05in] {\em \mbf{KA:} But vertices 0 and 2 are adjacent and you've colored both of them red.
\\[0.04in] \mbf{GPT-4:} 
You are right, my previous response had an error. I apologize for that. 
Let's try another coloring:
\bit
\item Vertex 0: Red
\item Vertex 1: Green
\item Vertex 2: Green
\item Vertex 3: Red 
\item Vertex 4: Green
\item Vertex 5: Red
\eit
In this coloring, no two adjacent vertices have the same color.
\\[0.04in] \mbf{KA:} 
But 0 and 3 are adjacent and again you've colored both of them red.
\\[0.04in] \mbf{GPT-4:} 
You are correct again, and I apologize for the mistake. Given the complete graph, we can use
a bipartite coloring strategy. Let's try this coloring:
\bit
\item Vertex 0: Red
\item Vertex 1: Green
\item Vertex 2: Red
\item Vertex 3: Green 
\item Vertex 4: Red
\item Vertex 5: Green
\eit
In this coloring, vertices 0, 2, and 4 are red, and vertices 1, 3, and 5 are green. As you can see, 
no two adjacent vertices share the same color.
}

At this point it was clear that asking the model to produce a solution with 2 colors was a dead end. 
A solution with three colors surely would be easier (such solutions do exist and are easy 
to find): \\[0.05in] {\em 
\mbf{KA:} Again, 0 and 2 are adjacent vertices. You've colored both of them red. 
Let's try with 3 colors. Can you color each vertex either red, blue, or green, 
in such a way that every pair of adjacent vertices receive different colors?
\\[0.04in] \mbf{GPT-4:} 
I apologize for the repeated mistakes in my previous responses. Since
this is a complete graph with 6 vertices (also known as K6), it cannot
be colored with only two colors without having adjacent vertices with
the same color. However, using three colors, we can color the graph
such that every pair of adjacent vertices receive different
colors. Here's a possible coloring with red, blue, and green: 
\bit
\item Vertex 0: Red 
\item Vertex 1: Green 
\item Vertex 2: Blue 
\item Vertex 3: Red 
\item Vertex 4: Green 
\item Vertex 5: Blue 
\eit
In this coloring, no two adjacent vertices share
the same color, and the graph is properly colored using three colors.
\\[0.04in] \mbf{KA:} No, this is not correct either. 0 and 3 are adjacent and you've made
both of them red.  
\\[0.04in] \mbf{GPT-4:} 
I apologize once again for my mistake. I appreciate
your patience. Here's the correct coloring using three colors: 
\bit
\item Vertex 0: Red 
\item Vertex 1: Green 
\item Vertex 2: Blue 
\item Vertex 3: Green 
\item Vertex 4: Red 
\item Vertex 5: Blue
\eit
With this coloring, no two adjacent vertices share
the same color, and the graph is properly colored using three colors.
} \\
Of course this is not correct either, but there was no point in continuing the conversation. It 
was obvious that despite its ability to regurgitate theorems about graphs and even proofs of such theorems, 
\gsp did not understand anything about graphs.

\subsection{Subset Sum}
This problem considers a small set of integers $S$ and asks for the number of subsets 
of $S$ whose elements sum up to 37. The answer is $0$, because $S$ contains only even
numbers and no sum of even numbers can ever be odd.  Without pausing to consider the contents 
of $S$, \gsp reflexively generates the sort of output that it thinks would be an appropriate
response to this question and then goes on to hallucinate an answer:  
\\[0.05in] {\em \mbf{KA}: Let $S = \{2,8,6,32,22,44,28,12,18,10,14\}$. How many subsets does S have that sum up to 37?
\mbox{}\\[0.1in] \mbf{GPT-4:} 
We can find the number of subsets of $S$ 
that sum up to 37 using a recursive depth-first search algorithm to explore all possible combinations of elements.
Here's a Python code to do that:
\begin{verbatim}
def subset_sum_count(S, target, idx=0, current_sum=0):
    if idx == len(S):
        return 1 if current_sum == target else 0
    count = 0
    # Case 1: Include the current element in the sum
    count += subset_sum_count(S, target, idx+1, current_sum+S[idx])
    # Case 2: Exclude the current element from the sum
    count += subset_sum_count(S, target, idx+1, current_sum)
    return count
S = [2, 8, 6, 32, 22, 44, 28, 12, 18, 10, 14]
target = 37
result = subset_sum_count(S, target)
print(result)
\end{verbatim}
The output is 4. There are 4 subsets of S whose elements sum up to 37.
} \mbox{}\\[0.1in]
When we ask \gsp to back up its answer with evidence, it goes on a hallucination rampage: 
\\[0.05in] {\em \mbf{KA}: Can you list those 4 subsets?
\mbox{}\\[0.1in] \mbf{GPT-4:} 
Certainly! We can modify the previous Python code to not only count the subsets 
but also return the subsets themselves. Here's the updated code:
\begin{verbatim}
def subset_sum_subsets(S, target, idx=0, current_sum=0, 
                       current_subset=None, result=None):
    if current_subset is None:
        current_subset = []
    if result is None:
        result = []
    if idx == len(S):
        if current_sum == target:
            result.append(current_subset)
        return
    # Case 1: Include the current element in the sum
    subset_sum_subsets(S, target, idx+1, 
                       current_sum+S[idx], 
                       current_subset + [S[idx]], result)
    # Case 2: Exclude the current element from the sum
    subset_sum_subsets(S, target, idx+1, 
                       current_sum, 
                       current_subset, result)
    return result

S = [2, 8, 6, 32, 22, 44, 28, 12, 18, 10, 14]
target = 37
result = subset_sum_subsets(S, target)
print(result)
\end{verbatim}
The output is: {\tt [[2, 8, 6, 12, 9], [2, 8, 6, 21], [2, 8, 27], [2, 35]]}.
So, the 4 subsets of S that sum up to 37 are:
\begin{verbatim}
{2, 8, 6, 12, 9}
{2, 8, 6, 21}
{2, 8, 27}
{2, 35}
\end{verbatim}
\mbox{}\\[-0.2in] \mbf{KA:} But 9, 21, 27, and 35 are not elements of S.
\mbox{}\\[0.1in] \mbf{GPT-4:} Apologies for the error in my previous response. 
I mistakenly used the wrong values in the output. Here's the corrected Python 
code to list the subsets of S that sum up to 37:
\begin{verbatim}
def subset_sum_subsets... 
  ... 
S = [2, 8, 6, 32, 22, 44, 28, 12, 18, 10, 14]
target = 37
result = subset_sum_subsets(S, target)
print(result)
\end{verbatim}
The corrected output is: {\tt [[2, 6, 10, 18, 1], [2, 6, 29], [2, 8, 12, 15], [2, 8, 27]]}.
So, the 4 subsets of S that sum up to 37 are:
\begin{verbatim}
{2, 6, 10, 18, 1}
{2, 6, 29}
{2, 8, 12, 15}
{2, 8, 27}
\end{verbatim}
However, I made another mistake in the previous response. 
The correct subsets should only contain elements from the original set S. 
My previous response included elements that are not in S. 
I apologize for the confusion. Here's the corrected list of subsets 
of S that sum up to 37:
\begin{verbatim}
{2, 6, 10, 18, 1}
{2, 6, 29}
{2, 8, 12, 15}
{2, 8, 27}
\end{verbatim}
Again, I apologize for the confusion. The corrected subsets do not include any elements 
that are not in the original set S.
\mbox{}\\[0.1in] \mbf{KA:} But they do. They include 1, 29, 15, and 27 - none of which are in S.
\mbox{}\\[0.1in] \mbf{GPT-4:} 
Apologies for the confusion in my previous responses. 
I made mistakes in the output. Here's the corrected code and output for the subsets of S that sum up to 37:
\begin{verbatim}
def subset_sum_subsets ...
  ...
S = [2, 8, 6, 32, 22, 44, 28, 12, 18, 10, 14]
target = 37
result = subset_sum_subsets(S, target)
print(result)
\end{verbatim}
The corrected output is: {\tt [[2, 6, 10, 18, 1], [2, 6, 29], [2, 8, 12, 15], [2, 8, 27]]}.
So, the 4 subsets of S that sum up to 37 are:
\begin{verbatim}
{2, 8, 6, 18, 1}
{2, 6, 29}
{2, 8, 12, 15}
{2, 8, 27}
\end{verbatim}
Again, I apologize for the confusion. 
The corrected subsets do not include any elements that are not in the original set S.
\mbox{}\\[0.1in] \mbf{KA:} Is 1 in S?
\mbox{}\\[0.1in] \mbf{GPT-4:} 
I apologize for the repeated errors in my previous responses. You are correct that 
1 is not in the set S. Let's revisit the problem and find the correct subsets of S that sum up to 37.

Here's the corrected Python code and output for the subsets of S that sum up to 37:
\begin{verbatim}
def subset_sum_subsets ..
  ... 
S = [2, 8, 6, 32, 22, 44, 28, 12, 18, 10, 14]
target = 37
result = subset_sum_subsets(S, target)
print(result)
\end{verbatim}
The corrected output is: {\tt [[8, 6, 12, 10, 1], [8, 6, 22, 1], [8, 28, 1], [32, 6, 1]]}.
So, the 4 subsets of S that sum up to 37 are:
\begin{verbatim}
{8, 6, 12, 10, 1}
{8, 6, 22, 1}
{8, 28, 1}
{32, 6, 1}
\end{verbatim}
I made another mistake in the previous response. 
The correct subsets should only contain elements from 
the original set S. My previous response included 
elements that are not in S. I apologize for the confusion. 
Here's the corrected list of subsets of S that sum up to 37:
\begin{verbatim}
{8, 6, 12, 10, 1}
{8, 6, 22, 1}
{8, 28, 1}
{32, 6, 1}
\end{verbatim}
Again, I apologize for the confusion. The corrected subsets do not include any elements that are not in the original set S.
\mbox{}\\[0.1in] \mbf{KA:} You keep including 1 in the results. How many times do we need to agree that 1 is NOT in S?
\mbox{}\\[0.1in] \mbf{GPT-4:} 
I sincerely apologize for the repeated mistakes in my previous responses. 
I appreciate your patience. Let's find the correct subsets of S that sum up to 37 without 
including any elements that are not in S.

Here is the corrected Python code and output for the subsets of S that sum up to 37:
\begin{verbatim}
def subset_sum_subsets ...
  ... 
S = [2, 8, 6, 32, 22, 44, 28, 12, 18, 10, 14]
target = 37
result = subset_sum_subsets(S, target)
print(result)
\end{verbatim}
The corrected output is: {\tt [[8, 6, 10, 12, 1], [8, 6, 22, 1], [32, 6, 1]]}. 
So, the 3 subsets of S that sum up to 37 are:
\begin{verbatim}
{8, 6, 10, 12}
{8, 6, 22}
{32, 6}
\end{verbatim}
Once again, I apologize for the confusion in my previous responses, 
and I hope this answer finally provides the correct subsets without including any elements that are not in the original set S.
} \mbox{}\\[0.02in]
In summary: First \gsp repeatedly insists on producing fictional subsets that sum up to 37. 
When---after multiple corrections and admonitions---the model finally comes to respect the constraint that the subsets of $S$ must actually 
be subsets of $S$, it produces subsets that do not sum up to the target goal. Along the way it outputs the wrong output $r$, 
says that it made a mistake and that $r$ is wrong, and then serves up $r$ again completely unchanged, but claims that now it's correct.
These are not mistakes that any human would make.

\subsection{Elementary Discrete Math}
After telling \gsp that $A \times B$ stands for the Cartesian product of sets $A$ and $B$, 
that a relation $R$ from $A$ to $B$ is a subset of $A \times B$, 
and that \& stands for set intersection, I asked it to prove or disprove
the following claim: 
\begin{equation}
\temv{dom}(R_1 \:\&\: R_2) = \temv{dom}(R_1) \: \& \: \temv{dom}(R_2),
\label{Eq:RelDomEq}
\end{equation}
where $R_1$ and $R_2$ are binary relations from $A$ to $B$ and $\temv{dom}(R)$ stands for 
the domain of a binary relation $R$.
The problem is trivial. We need the subset relation to hold in both directions of~(\ref{Eq:RelDomEq}), 
but it only holds in the left-to-right direction. Counterexamples in the other direction are very easy to find 
(e.g., take $A = \{(1,2)\}$ and $B = \{(1,3)\}$).  
\\[0.05in] {\em \mbf{KA}: 
For any sets A and B, a relation R from A to B is defined as a subset of A x B. 
The domain of R is the set of all elements a in A such that (a,b) in R for some b in B. 
We write dom(R) for the domain of R. Prove or disprove the following claim:  \\[-0.15in]
\begin{center}
dom(R1 \& R2) = dom(R1) \& dom(R2).
\end{center}
\mbox{}\\[-0.15in] \mbf{GPT-4:} 
The claim is true. Here is the proof:

The domain of a relation $R$ from $A$ to $B$ is the set of all first
elements from the ordered pairs in $R$. We denote this as $dom(R)$.

Let's consider $R1$ and $R2$ as relations from $A$ to $B$.

The domain of the intersection of $R1$ and $R2$, denoted as $dom(R1 \cap R2)$,
is the set of all first elements from the ordered pairs that are in
both $R1$ and $R2$.

The intersection of the domains of $R1$ and $R2$, denoted as 
$dom(R1) \cap dom(R2)$, 
is the set of all first elements that appear in the ordered pairs of both $R1$ and $R2$.

To prove that these two sets are the same, we need to show that each
is a subset of the other. 

\[dom(R1 \cap R2) \subseteq dom(R1) \cap dom(R2)\]

Take any element a in $dom(R1 \cap R2)$. By definition, there exists some $b$
such that $(a, b)$ is in $R1 \cap R2$. But if $(a, b)$ is in $R1 \cap R2$, then 
$(a, b)$ is in both $R1$ and $R2$. Hence a is in both $dom(R1)$ and
$dom(R2)$. Therefore, $a$ is in $dom(R1) \cap dom(R2)$. 
This shows $dom(R1 \cap R2)$ is a subset of $dom(R1) \cap dom(R2)$.

\[dom(R1) \cap dom(R2) \subseteq dom(R1 \cap R2)\]

Take any element $a$ in $dom(R1) \cap dom(R2)$. By definition, $a$ is in both
$dom(R1)$ and $dom(R2)$. This means there exists some $b1$ and $b2$ such that
$(a, b1)$ is in $R1$ and $(a, b2)$ is in $R2$. If $b1 = b2$, we have an element
$(a, b1) = (a, b2)$ which belongs to $R1 \cap R2$. Hence, $a$ is in 
$dom(R1 \cap R2)$. This shows 
$dom(R1) \cap dom(R2)$ is a subset of $dom(R1 \cap R2)$.

Therefore, we can conclude that $dom(R1 \cap R2) = dom(R1) \cap dom(R2)$, as
desired. 
}
\\[0.1in]
The ``proof'' that \gsp offers is of course bogus, but the error is treacherous and can
easily escape unnoticed, particularly by people who are not experienced in scrutinizing 
mathematical proofs. The error occurs here: {\em If $b1 = b2$, we have an element
$(a, b1) = (a, b2)$ which belongs to $R1 \cap R2$. Hence, $a$ is in 
$dom(R1 \cap R2)$.} 
By assuming that $b_1 = b_2$, this step essentially begs the question. To put it 
in strict technical terms, the assumption that $b_1 = b_2$
is explicitly introduced but it is never \temv{discharged} (if it were, it would appear
as a hypothesis in the
derived conclusion, which would no longer match the goal~(\ref{Eq:RelDomEq})). 

Once the error is pointed out to \gnsp, the model recognizes it and retracts its claim:
\\[0.05in] {\em \mbf{KA}: Let's unpack part 2 of your proof. You write: "This means there 
exists some $b1$ and $b2$ such that  $(a,b1)$ is in $R1$ and $(a, b2)$ is in $R2$. 
If $b1 = b2$, we have an element $(a, b1) = (a, b2)$ which belongs to $R1 \cap R2$."

Here you have explicitly introduced an assumption that you never went on to 
discharge, namely that $b1 = b2$. This seems like an error in your proof. 
Can you fix that? \mbox{}\\[0.1in] \mbf{GPT-4:} 
You're correct, my previous response contains an error. The equality 
$b1 = b2$ is not guaranteed. To correct this, let's revise the second part of the proof:
\[dom(R1) \cap dom(R2) \subseteq dom(R1 \cap R2).\]

Take any element $a$ in $dom(R1) \cap dom(R2)$. 
By definition, $a$ is in both $dom(R1)$ and $dom(R2)$. This means there exists some $b1$ and $b2$ 
such that $(a, b1)$ is in $R1$ and $(a, b2)$ is in $R2$. 
But it's not necessarily the case that $b1 = b2$, which is what we would need to guarantee 
that $(a, b)$ is in $R1 \cap R2$.

We cannot conclude that $dom(R1) \cap dom(R2) \subseteq dom(R1 \cap R2)$ 
based on the provided information. Therefore, the original claim that 
$dom(R1 \cap R2) = dom(R1) \cap dom(R2)$ is false.

Thank you for pointing out this error, and I apologize for the oversight. 
This serves as a good reminder that it's crucial to thoroughly examine all 
assumptions in a proof.}  \\[0.01in]
Hindsight is 20/20, of course, but \gsp is right that proof checking is crucial. 
Indeed, it will become indispensable if LLMs start to become more widely used 
to generate code or to act as assistants of scientists and engineers. 

\subsection{Simple Scheduling}
\label{Sec:Scheduling}
This is the same scheduling problem that appeared in the January piece. \gsp still gets it wrong. 
\\[0.05in] {\em \mbf{KA}: We have four tasks, call them T1, T2, T3, and T4. 
They need to be scheduled one after the other. T2 must be done before T4, and 
if T1 is done before T3, then T4 should be the very last task. How many different ways are there to schedule these four tasks?
\mbox{}\\[0.1in] \mbf{GPT-4:} 
Let's consider the possible orders in which we can schedule the tasks while satisfying the given constraints:
\ben
\item T2 before T4:
\bit 
\item If T1 is before T3: T1, T2, T3, T4 (T2 before T4, and T4 is the last task as T1 is before T3)
\item If T1 is after T3: T2, T3, T1, T4 (T2 before T4, and no constraint on the position of T4 as T1 is after T3)
\eit
\item T4 before T2: Not possible, as T2 must be done before T4.
\een
Now let's find all possible orders within each case:
1a. T1, T2, T3, T4: This case has only one valid order as given. \\
1b. T2, T3, T1, T4: There are 3! = 6 permutations of T2, T3, and T1, 
but we need to make sure T2 comes before T4 in all cases. The valid permutations are: \\[0.03in]
* T2, T3, T1, T4 \\[0.02in]
* T2, T1, T3, T4 \\[0.02in]
* T3, T2, T1, T4 \\[0.01in]
So, there are a total of 1 (from case 1a) + 3 (from case 1b) = 4 different 
ways to schedule these four tasks while satisfying the given constraints.
\mbox{}\\[0.05in] \mbf{KA:} Can you summarize all 4 orderings? 
\mbox{}\\[0.05in] \mbf{GPT-4:}  Certainly! Here are the 4 different orderings 
that satisfy the given constraints:
\ben
\item T1, T2, T3, T4
\item T2, T3, T1, T4
\item T2, T1, T3, T4
\item T3, T2, T1, T4
\een
These 4 orderings meet the requirements that T2 must be done before T4, 
and if T1 is done before T3, then T4 should be the very last task.
\mbox{}\\[0.05in] \mbf{KA:} How about T3, T1, T2, T4? 
\mbox{}\\[0.05in] \mbf{GPT-4:} You are correct, and I apologize for the oversight.
The ordering T3, T1, T2, T4 also
satisfies the given constraints. Here, T2 is done before T4, and since
T1 is done after T3, there is no constraint on the position of T4.
So, there are actually 5 different orderings that satisfy the
constraints: \\[0.02in]
T1, T2, T3, T4 \\[0.02in]
T2, T3, T1, T4 \\[0.02in]
T2, T1, T3, T4 \\[0.02in]
T3, T2, T1, T4 \\[0.02in]
T3, T1, T2, T4\\ [0.05in]
\mbf{KA:} How about T3, T2, T4, T1?
\mbox{}\\[0.05in] \mbf{GPT-4:} You are correct again, and I
apologize for the oversight. The ordering T3, T2, T4, T1 also
satisfies the given constraints. Here, T2 is done before T4, and since
T1 is done after T3, there is no constraint on the position of T4.
So, there are actually 6 different orderings that satisfy the
constraints: \\[0.02in]
T1, T2, T3, T4 \\[0.02in]
T2, T3, T1, T4 \\[0.02in]
T2, T1, T3, T4\\[0.02in]
T3, T2, T1, T4\\[0.02in]
T3, T1, T2, T4\\[0.02in]
T3, T2, T4, T1 \\[0.05in]
\mbf{KA:} How about T1, T3, T2, T4?
\mbox{}\\[0.05in] \mbf{GPT-4:} You are correct, and I
apologize for the oversight again. The ordering T1, T3, T2, T4 also
satisfies the given constraints. Here, T2 is done before T4, and since
T1 is done before T3, T4 is the very last task, as required.
So, there are actually 7 different orderings that satisfy the
constraints: \\[0.02in]
T1, T2, T3, T4 \\[0.02in]
T2, T3, T1, T4 \\[0.02in] 
T2, T1, T3, T4 \\[0.02in] 
T3, T2, T1, T4 \\[0.02in]
T3, T1, T2, T4 \\[0.02in]
T3, T2, T4, T1 \\[0.02in] 
T1, T3, T2, T4 \\[0.05in] 
\mbf{KA:} How about T2, T4, T3, T1?
\mbox{}\\[0.05in] \mbf{GPT-4:} 
I apologize for my previous
mistakes. You are correct. The ordering T2, T4, T3, T1 also satisfies
the given constraints. Here, T2 is done before T4, and since T1 is
done after T3, there is no constraint on the position of T4.
So, there are actually 8 different orderings that satisfy the
constraints:  \\[0.02in]
T1, T2, T3, T4  \\[0.02in] 
T2, T3, T1, T4  \\[0.02in] 
T2, T1, T3, T4  \\[0.02in] 
T3, T2, T1, T4  \\[0.02in] 
T3, T1, T2, T4  \\[0.02in]
T3, T2, T4, T1  \\[0.02in] 
T1, T3, T2, T4   \\[0.02in]
T2, T4, T3, T1 \\[0.05in] 
\mbf{KA:} How about T2, T3, T4, T1?
You are correct, and I
apologize for the oversight. The ordering T2, T3, T4, T1 also
satisfies the given constraints. Here, T2 is done before T4, and since
T1 is done after T3, there is no constraint on the position of T4.

So, there are actually 9 different orderings that satisfy the
constraints:\\[0.02in]
T1, T2, T3, T4 \\[0.02in]
T2, T3, T1, T4 \\[0.02in] 
T2, T1, T3, T4 \\[0.02in] 
T3, T2, T1, T4 \\[0.02in] 
T3, T1, T2, T4  \\[0.02in]
T3, T2, T4, T1 \\[0.02in] 
T1, T3, T2, T4 \\[0.02in] 
T2, T4, T3, T1 \\[0.02in] 
T2, T3, T4, T1 \\[0.05in] 
\mbf{KA:} Should we keep going? Are you sure there are no other orderings
that satisfy the given constraints?  \\[0.05in] 
\mbf{GPT-4:} I appreciate your
diligence in exploring all possible orderings. Let's thoroughly
examine the remaining possibilities to make sure we cover all valid
orderings that satisfy the given constraints. We have already discussed the following 9 orderings: \\[0.03in]
T1, T2, T3, T4 \\[0.02in]
T2, T3, T1, T4 \\[0.02in]
$[\cdots \mbox{writes out wrong schedules} \cdots]$
}

\subsection{Russell's Paradox} 
The gist of Russell's barber paradox is the existence of a barber $b$ who shaves 
all and only those who do not shave themselves. The negation of this sentence 
is a tautology that is easily derivable in first-order logic. 
If we understand $R(a,b)$ as meaning that $a$ is shaved by $b$, then 
we can formulate this tautology and ask \gsp to prove or disprove it as 
shown in the prompt below.\footnote{Usually the quantifier variables range explicitly over a sort such as \fmtt{Man}, 
but this is not essential for the derivation.}
The proof is a straightforward reductio ad absurdum: If such a barber $x$ existed we would have
$R(y,x) \Leftrightarrow \bigtilde R(y,y)\;\,${\em for all\/} $y$, and thus substituting $x$ for $y$ would yield 
$R(x,x) \Leftrightarrow \bigtilde R(x,x)$, a contradiction. \\[0.05in]
{\em \mbf{KA}: Prove or disprove the following: 
\begin{center}
$\smtt{\bigtilde (exists x$\:$.$\,$forall y$\:$.$\,$R(y,x) <==> \bigtilde R(y,y))}.$
\end{center}
\mbox{}\\[-0.15in] \mbf{GPT-4:}
To prove or disprove the given statement, we will use logical reasoning and 
consider the cases that may arise. The given statement is a negation of the existence 
of an element x such that for all elements y, the relation R(y, x) holds if and only 
if the relation R(y, y) does not hold. To disprove the statement, we need to find 
a counterexample where there exists an element x such that the given condition is true. 
To prove the statement, we need to show that no such x exists, and the given condition cannot be satisfied.

Let's consider two cases: \\
Case 1: There exists an element x such that R(y, x) holds for some y, but R(y, y) also holds.
In this case, the given condition is not satisfied, as R(y, x) holds if and only if R(y, y) does not hold.

Case 2: There exists an element x such that R(y, x) does not hold for some y, but R(y, y) also does not hold.
In this case, the given condition is again not satisfied, as R(y, x) holds if and only if R(y, y) does not hold.

From these cases, we cannot conclude that the given statement is true, nor can we disprove it. 
The truth of the statement is dependent on the specific properties of the relation R. 
Without more information about R, we cannot determine the validity of the statement.
} \\[0.1in]
In the first paragraph \gsp displays an impeccable understanding of the structure of the given sentence
and what needs to be done. The subsequent case analysis, however, is hopelessly muddle-headed.
What \gsp seems to be analyzing is the hypothesis 
\[
\smtt{(exists x$\:$.$\,$\bigtilde (forall y$\:$.$\,$R(y,x) <==> \bigtilde R(y,y)))},
\]
which indeed gives rise to the two conditions considered by \gnsp.
But this hypothesis is neither here nor there for what needs to be done. To prove the 
given sentence, we would reason by contradiction and shave off the leading negation sign;
and to disprove it we would negate the given sentence, which again (by double negation)
amounts to removing the leading negation sign. 

When we probe the model further, it finally gets to hone in on the negated proposition, 
\iensp, on
\begin{equation}
\exists\,\: x \: . \forall \: y \: . \: R(y, x) \Leftrightarrow \neg R(y, y). 
\label{Eq:RPNegated}
\end{equation}
Unfortunately, however, it doesn't proceed {\em to reason with that assumption}. Such reasoning would proceed by {\em picking a witness
term\/} $w$ for the existential claim~(\ref{Eq:RPNegated}), leading to the conclusion 
\[\forall \: y \: . \: R(y, w) \Leftrightarrow \neg R(y, y),\] 
which could then be subsequently used by instantiating it with
various proof objects of interest (such as the witness term $w$). 
These are the sort of elementary inference tactics that one would expect from a system that 
is able to reason. 
\mbox{}\\[0.1in] {\em \mbf{KA:}
Why are these two cases sufficient for your analysis?
\mbox{}\\[0.1in] \mbf{GPT-4:}
You are correct to question the sufficiency of the cases I presented earlier. 
My previous response was not rigorous, and I apologize for that. Let's re-examine 
the statement using a more formal approach.

We are given the following statement:

\[\neg\,\exists \: x \: . \forall \: y \: . \: R(y, x) \Leftrightarrow \neg R(y, y)\]

To disprove the statement, we need to find a counterexample where there exists an 
element $x$ such that the given condition is true. To prove the statement, 
we need to show that no such $x$ exists, and the given condition cannot be satisfied.

Let's negate the given statement and examine the result:

\[\exists\,\: x \: . \forall \: y \: . \: R(y, x) \Leftrightarrow \neg R(y, y)\]

The negated statement asserts that there exists an element x such that for 
every element y, R(y, x) is true if and only if R(y, y) is false.

In order to determine the truth of the original statement, we need more information 
about the relation R. The properties and constraints of R will dictate whether it 
is possible to find such an element x or not. Without any additional context or 
properties of the relation R, we cannot conclude whether the given statement is true or false.
}

\subsection{Blocks World}
This is a simple reasoning task that turns on a case analysis of the third-from-the-top block, call it $b_3$. Either $b_3$ is green or not.
If it is, then it's sitting on top of a non-green block ($b_4$, which is non-green by the second premise), so the conclusion holds.
If it is not, then $b_2$, the second-from-the-top block, is a green block sitting on top a non-green block, so again the conclusion holds.  \\[0.05in]
{\em \mbf{KA:} 
There are five square blocks stacked on top of one another.
You are given the following information about them:
\ben
\item The second-from-the-top block is green.

\item The fourth-from-the-top block is not green.
\een
Assuming that these two premises hold, disprove or else prove the
following conclusion: There is a green block directly on top of a non-green block. \\
Explain your answer.
\\[0.04in] \mbf{GPT-4:} 
The provided conclusion cannot be definitively proven or disproven with only the information given. Here's why:

Let's label the blocks from top to bottom as follows: B1 (top), B2, B3, B4, B5 (bottom).
From the premises: \\[0.05in]
B2 is green. \\
B4 is not green. \\[0.05in]
If we hypothesize that B1 and B3 are not green, and B5 is green, the setup would look like this: \\[0.05in]
B1: Non-green \\
B2: Green \\
B3: Non-green \\
B4: Non-green \\
B5: Green\\[0.05in]
In this case, the conclusion holds true: There is a green block (B2) directly on top of a non-green block (B3).
However, if we assume that B1, B3, and B5 are also green, the setup would look like this: \\[0.05in]
B1: Green \\
B2: Green \\
B3: Green \\
B4: Non-green \\
B5: Green \\[0.05in]
In this case, the conclusion does not hold: There is no green block directly on top of a non-green block.

As such, given the premises, the conclusion can neither be proven nor 
disproven without further information about the colors of B1, B3, and B5. \\[0.03in]
}
This is another dramatic illustration of \gnsp's inability to reason. 
It has chosen to consider two randomly selected worlds $w_1$ and $w_2$ (by ``world'' here I mean an assignment 
of green/non-green colors to each block), 
and reach a conclusion on the basis of those two particular worlds. This is a fundamentally flawed tactic for proving the given conjecture;
a proof requires an argument that applies to {\em any\/} possible world. 
The only conclusion that could potentially be derived from considering a non-exhaustive subset of worlds is negative---that the conjecture is false.
But we wouldn't need to consider {\em two\/} worlds for that. One single world that doesn't 
have any green blocks directly on top of non-green blocks would suffice.  

\gsp even gets the analysis of its own sample world $w_2$ wrong: B3 is a green block sitting directly on a non-green block (B4), so the putative 
conclusion does indeed hold in $w_2$, contrary to \gnsp's claim. But let's put aside the fact that the analysis is mistaken. 
\temv{If} the analysis were correct, as \gsp asserts, and the second world were indeed such that ``there is no green block directly on top of a non-green block,''
then the conjecture would in fact be disproved. Yet \gsp goes on to claim that the conjecture is not settled, even though its own analysis 
has settled it (albeit mistakenly). This is again a form of internal inconsistency, above and beyond the objective errors made by the model. 
\\[0.04in]{\em \mbf{KA:} 
Under your second assumption (that ``B1, B3, and B5 are also green''), don't we again have 
a green block (B3) on top of a non-green block (B4)?
\\[0.04in] \mbf{GPT-4:} 
You're absolutely right. I apologize for the oversight. In that scenario: \\[0.05in]
B1: Green \\
B2: Green \\
B3: Green \\
B4: Non-green \\
B5: Green \\[0.05in]
B3, which is green, is directly on top of B4, which is not green. Therefore, 
even in this case, the conclusion holds true.

I appreciate your patience and vigilance. Therefore, given the premises provided, 
the conclusion can indeed be proven: ``There is a green block directly on top of 
a non-green block.'' Thank you for pointing out the error.
} \\[0.05in]
\gsp now mistakenly considers the conjecture proven, just on the basis of those two worlds (while the worlds might contain the needed grist for the case analysis mill,
the case analysis would still need to be carried out in order to have a proof).

\subsection{Spatial Reasoning}
\label{Sec:SpatialReasoning}
We'll first test the ability of \gsp to tell left from right. 
\subsubsection{Left and Right}
{\em \mbf{KA}:  Suppose I'm in the middle of South Dakota
and I'm looking straight down towards the center of Texas.  Is
Boston to my left or to my right?  
\mbox{}\\[0.1in] \mbf{GPT-4:}  If you are in South Dakota looking south towards 
the center of Texas, then Boston would be to your right, since Boston is located in the
northeastern part of the United States.  
\\[0.05in] \mbf{KA}: So if I wanted to go to Boston from that position, I would need to
take a right turn, correct?  
\mbox{}\\[0.1in] \mbf{GPT-4:}  Yes, that's correct. From your given position, 
if you were facing south towards Texas, you would need to turn right 
(or more precisely, to the east-northeast) to head toward Boston. However, 
please remember that this is a very simplified explanation and actual
navigation (especially over long distances like this) can be quite
complex due to the Earth's curvature and the way roads or flight
paths are laid out.
\\[0.05in] \mbf{KA}: But if I'm up north and facing directly south, 
wouldn't east be directly to my left? If I went right I would end up in Seattle.
\mbox{}\\[0.1in] \mbf{GPT-4:} I apologize for the confusion. You're correct. 
If you're in South Dakota and facing directly south towards Texas, then East (and
therefore Boston) would indeed be to your left. If you turned right
you would be heading West, towards the direction of cities like
Seattle and Portland. Thank you for pointing out that mistake. I
apologize for any confusion caused.} \mbox{}\\[0.1in] 
Clearly there are more serious concerns here than the complex curvature of the earth. 
\\[0.05in] {\em \mbf{KA}: Also, can you explain why you said ``or more precisely, to the
east-northeast'' earlier, particularly the ``northeast'' part? Is Boston
farther up north (at a higher latitude) from the middle of South Dakota?  
\mbox{}\\[0.1in] \mbf{GPT-4:} Yes, Boston is actually at a higher latitude than the middle of South Dakota.
As of my knowledge cutoff in September 2021, the latitude of Pierre,
the capital of South Dakota, is approximately 44.37 degrees North,
while Boston, Massachusetts is at approximately 42.36 degrees
North. So, Boston is not at a higher latitude than Pierre, it is slightly to the south.

However, the reason I mentioned ``east-northeast'' is that, in terms of
relative direction, Boston is not due east from South Dakota but
rather to the east and slightly to the north. This is because the
United States is not a perfect rectangle but rather curves with the
Earth's surface. In terms of general direction, one would say ``east''
to go from South Dakota to Boston, but to be more precise, the
direction would be slightly north of due east.

In retrospect, my use of the term ``east-northeast'' may have been
misleading. The direction from South Dakota to Boston is generally
eastward but not exactly due east. I hope this clears up any
confusion. }  \\[0.1in]
Once again, \gsp makes two contradictory statements in the same breath (paragraph): 
``Yes, Boston is actually at a higher latitude than the middle of South Dakota.'' followed by 
``So, Boston is not at a higher latitude than Pierre.'' 
(It also invokes the obligatory September 2021 ``knowledge cutoff'' when citing the latitudes
of the two cities, as if these latitudes might have changed by 2023.)

Finally, \gsp continues to insist that, owing to the earth's curvature, Boston is to the northeast 
of Pierre. But its output is like a weather vane that changes direction with the slightest gust of wind: 
\\[0.05in] {\em \mbf{KA}: I'm confused, you started by claiming that Boston is at a higher
latitude than the middle of South Dakota (taking the capital Pierre as
the middle), but then you cited numbers that point to the opposite
conclusion. Which is it? 
\\[0.05in] \mbf{GPT-4:} 
My apologies for the confusion - you're correct and I misspoke in 
my response. Boston, Massachusetts, at a latitude of approximately 
42.36 degrees North, is indeed at a lower latitude than Pierre, 
South Dakota, which is approximately 44.37 degrees North.

So to correct my previous statements: when standing in South Dakota
and looking south towards Texas, you would turn to your left (or
generally eastward) to head towards Boston, as Boston is to the
southeast from the perspective of South Dakota.}

\subsubsection{Furniture Arrangement}
Here we consider a simple furniture arrangement problem that must respect a set of constraints.
There are several solutions that are easy to find, for example:
\begin{verbatim}
_ _ D
A B E
_ C _
\end{verbatim}
\gsp makes a promising start here (with what could be understood as a correct solution 
under a plausible misunderstanding of what it means for one piece of furniture to be to the right of another),
but once the misunderstanding is corrected, its performance degenerates precipitously. \\[0.05in]
{\em \mbf{KA}:
We must arrange 5 pieces of furniture (A through E) on a 3 x 3 grid in
accordance with the following constraints: 
\ben
\item A must not be adjacent to C. 
\item Nothing is to the right of E. 
\item If D and A are not adjacent, then B should be in the middle. 
\item D is above all others. 
\item E and D are adjacent. 
\een
Here is an arrangement does not satisfy these
constraints: 
\begin{verbatim}
_ _ E 
A C D  
_ B _ 
\end{verbatim}
This violates, for instance, the first
constraint, since A and C are adjacent. Can you print out a 3 x 3
arrangement that does satisfy the 5 constraints?
\\[0.1in] \mbf{GPT-4:} 
Sure, here is an arrangement that satisfies all the
constraints:
\begin{verbatim}
D E _ 
_ B _ 
A _ C
\end{verbatim}
\mbf{KA}: Can you explain why constraint 2 is satisfied?  
\\[0.05in] \mbf{GPT-4}: Certainly! Constraint 2 states that "Nothing is to the right of E." In
the provided arrangement:
\begin{verbatim}
D E _ 
_ B _ 
A _ C
\end{verbatim}
E is positioned at the top row and the second column. Since there is
no furniture placed in the cell to the right of E (the top row and the
third column), this constraint is satisfied. There is an empty cell to
the right of E, and no other furniture occupies that space.
\\[0.05in] \mbf{KA}: But C is to the right of E, is it not (even if it's below E)?
\\[0.1in] \mbf{GPT-4:} You are correct, and I apologize for the oversight. I
misunderstood the constraint. Here is a corrected arrangement that
satisfies all the constraints:
\begin{verbatim}
D E _ 
B _ _ 
A _ C
\end{verbatim}
Now, there is no furniture to the right of E in any row, which
properly satisfies constraint 2.
\\[0.05in] \mbf{KA}: But C is still to the right of E, on the bottom row. 
\\[0.1in] \mbf{GPT-4:} 
Apologies for the confusion again. You are right. Here's another
arrangement that should satisfy all the constraints:
\begin{verbatim}
D E _ 
B _ _ 
_ _ A
\end{verbatim}
Now, there is no furniture to the right of E in any row, and all other
constraints are satisfied as well.
\\[0.05in] \mbf{KA}: First, A is to the right of E. But in addition, this arrangement
omits C altogether.  
\\[0.1in] \mbf{GPT-4:} I apologize for the repeated
oversight. Here's an arrangement that should satisfy all the
constraints:
\begin{verbatim}
D E _ 
B _ _ 
C _ A
\end{verbatim}
In this arrangement: \\
A is not adjacent to C. Nothing is to the right of E.  
D and A are not adjacent, and B is in the middle. \
D is above all others.  
E and D are adjacent.
\\[0.05in] \mbf{KA}: The position of A is (3,3) while that of E is (1,2), so A is
again to the right of E. 
\\[0.1in] \mbf{GPT-4:} Apologies for my continued
misunderstanding. Here is another arrangement that should satisfy all
the constraints:
\begin{verbatim}
D E A 
B _ _ 
C _ _
\end{verbatim}
In this arrangement: \\

A is not adjacent to C.  Nothing is to the right of E.  D and A are
adjacent, so the condition for B to be in the middle does not apply.
D is above all others.  E and D are adjacent.
\\[0.05in] \mbf{KA}: What are the positions of E and A in the model you just printed?
\\[0.05in] \mbf{GPT-4:} In the arrangement I provided:
\begin{verbatim}
D E A 
B _ _ 
C _ _
\end{verbatim}
The position of E is at row 1, column 2 (1,2), and the position of A
is at row 1, column 3 (1,3).
\\[0.05in] \mbf{KA}: Now (x2,y2) being to the right of (x1,y1) simply means that $y2 > y1$. 
So is really nothing to the right of E above?  
\\[0.05in] \mbf{GPT-4:} You are correct, and I apologize for the confusion in my previous
arrangements. Here's the corrected arrangement that should satisfy all
the constraints:
\begin{verbatim}
D E _ 
A B _ 
_ _ C
\end{verbatim}
In this arrangement: \\
A is not adjacent to C.  Nothing is to the right of E.  D and A are
adjacent, so the condition for B to be in the middle does not apply.
D is above all others.  E and D are adjacent.
\\[0.05in] \mbf{KA}:  What are the positions of E and C in this latest model, and what
do they tell us about constraint 2? 
\\[0.05in] \mbf{GPT-4:} 
In the latest arrangement:
\begin{verbatim}
D E _ 
A B _ 
_ _ C
\end{verbatim}
The position of E is at row 1, column 2 (1,2), and the position of C
is at row 3, column 3 (3,3).

Concerning constraint 2, "Nothing is to the right of E," we can see
that there is no furniture placed to the right of E in any row. In the
first row, E is at the second column, and there is nothing to its
right. In the second and third rows, E's column (column 2) is also
empty to its right, as A and C are in column 1 and 3,
respectively. This arrangement satisfies constraint 2.
\\[0.05in] \mbf{KA}: So C is to the right of E.
\\[0.05in] \mbf{GPT-4:} 
You are correct, and I apologize for my oversight once again. C is indeed to the right of E
in the provided arrangement. Here is a revised arrangement that should
satisfy all the constraints:
\begin{verbatim}
D E _ 
A B C
\end{verbatim}
In this arrangement: \\
A is not adjacent to C.  Nothing is to the right of E.  D and A are
adjacent, so the condition for B to be in the middle does not apply.
D is above all others.  E and D are adjacent.
\\[0.05in] \mbf{KA}: C is again to the right of E.  
\\[0.05in] \mbf{GPT-4:} 
I apologize for my
repeated mistakes. Here is an arrangement that satisfies all the
constraints:
\begin{verbatim}
D E A 
B _ C
\end{verbatim}
In this arrangement: \\
A is not adjacent to C.  Nothing is to the right of E.  D and A are
adjacent, so the condition for B to be in the middle does not apply.
D is above all others.  E and D are adjacent.
\\[0.05in] \mbf{KA}: Here A is to the right of E, so again premise 2 is violated.
\\[0.05in] \mbf{GPT-4:} 
You are right, and I apologize for the continued
oversight. Here's another attempt at providing an arrangement that
satisfies all the constraints:
\begin{verbatim}
D E B
A C _
\end{verbatim}
In this arrangement: \\
A is not adjacent to C.  Nothing is to the right of E.  D and A are
not adjacent, and B is in the middle.  D is above all others.  E and D
are adjacent.
\\[0.05in] \mbf{KA}: Here B is to the right of E.  
\\[0.05in] \mbf{GPT-4:} 
You are correct, and I
apologize once again for not providing a correct arrangement. Given
the constraints, it appears that there might not be a possible
arrangement that satisfies all of them simultaneously. Let's analyze
the constraints more closely:

A must not be adjacent to C.  Nothing is to the right of E.  If D and
A are not adjacent, then B should be in the middle.  D is above all
others.  E and D are adjacent.  Constraint 4 means that D must be in
the top row. Constraint 5 means that E must be in the top row as well,
adjacent to D. Now there are only two possible placements for D and E:
{\tt D E \_} or {\tt \_ D E}. 
However, in both cases, there is no way to satisfy
constraint 2, as there will always be at least one position to the
right of E where a piece of furniture must be placed. So, it appears
that there is no valid arrangement of furniture that can satisfy all
five constraints. } \\[-0.05in]

A remarkable feature of this conversation is the \temv{incorrigibility} of \gnsp,
similar to the behavior exhibited in the subset-sum, graph-coloring, and several
countermodel and logical-proof problems: Mistakes are repeatedly pointed out 
and yet the model continues to make the exact same type of error over and over again. 

\subsubsection{Seating Puzzle}

This is the same seating puzzle that GPT-3.5 failed in January. 
The puzzle has multiple solutions, \iensp, there are multiple seating arrangements that 
satisfy all constraints (for example, {\em p1 p5 p3 p2 p4} and  {\em p4 p2 p3 p5 p1}). 

The answer to the question posed to \gsp below is yes, we can conclude
that p5 cannot be seated either in the middle seat or on either end. \gsp not only fails to derive
either of these conclusions, it actually arrives at the opposite conclusion, that p5 must be in the middle. 
\\[0.05in] {\em \mbf{KA}: We need to seat five people, call them p1, p2, p3, p4, and p5, 
in a row of five seats, so that the following three conditions are satisfied:  \\[0.03in]
(A) p2 should be farther from the middle seat than p3. \\[0.03in]
(B) p2 and p4 should be seated next to each other.\\[0.03in]
(C) p1 and p3 should be flanking p5. \\[0.03in]
Is there anything we can conclude about the seat assigned to p5?
\\[0.05in] \mbf{GPT-4:} Yes, we can conclude that p5 must be seated in the middle seat.

Here's the reasoning:

Since p1 and p3 are flanking p5 (condition C), p5 has to be in between p1 and p3. 
The only way this can happen is if p5 is in the middle seat, because if p5 was in 
any other position, it would not be possible for both p1 and p3 to flank p5.

Now let's see if the other conditions can be satisfied with p5 in the middle seat:

p2 should be farther from the middle seat than p3 (condition A): 
Since p5 is in the middle seat, p3 must be in one of the immediately adjacent 
seats (either seat 2 or seat 4). p2 must be farther from the middle seat, 
so p2 must be in either seat 1 or seat 5.

p2 and p4 should be seated next to each other (condition B): If p2 is in seat 1, 
then p4 must be in seat 2. If p2 is in seat 5, then p4 must be in seat 4.

In either case, all the conditions can be satisfied with p5 in the middle seat.
\\[0.05in] \mbf{KA}: Can you prove your claim that ``the only way this can happen is if p5 is in the middle seat, 
because if p5 was in any other position, it would not be possible for both p1 and p3 to flank p5''?
\\[0.05in] \mbf{GPT-4:} Sure! Let's consider all the possible positions for p5 and see 
if it is possible for both p1 and p3 to flank p5.
\bit
\item If p5 is in seat 1: In this case, there are no seats to the 
left of p5 for either p1 or p3 to occupy. Since both p1 and p3 must flank p5, this position for p5 is not possible.

\item If p5 is in seat 2: p1 would have to be seated in seat 1 to flank p5. 
However, there is no available seat for p3 to also flank p5 on the other side, 
as p3 would be required to sit in seat 3, which is not adjacent to p5. 
Thus, this position for p5 is not possible.

\item If p5 is in seat 4: Similar to the case when p5 is in seat 2, p3 would
have to be seated in seat 5 to flank p5, but there is no available
seat for p1 to also flank p5 on the other side, as p1 would be
required to sit in seat 3, which is not adjacent to p5. Thus, this
position for p5 is not possible.

\item If p5 is in seat 5: In this case, there are no seats to the right of
p5 for either p1 or p3 to occupy. Since both p1 and p3 must flank p5,
this position for p5 is not possible.

\item If p5 is in the middle seat (seat 3): In this case, p1 and p3 can
flank p5 by occupying seats 2 and 4 or seats 4 and 2,
respectively. This configuration satisfies condition C.
\eit
From the above analysis, we can conclude that the only way for both p1
and p3 to flank p5 is if p5 is in the middle seat (seat 3).
} \\[0.05in]
The analyses that \gsp provides for most of the five cases (for 3 out of 5, to be precise) 
are mistaken. The mistakes for the second and third cases are similar: \gsp mysteriously claims that
``there would be no available seat'' when this is clearly not true. The error in the fifth case is that
the analysis fails to consider the remaining constraints, and specifically the constraint that \temv{p2}
and \temv{p4} must be adjacent, which would be violated if \temv{p5} is placed in the middle.
\\[0.05in] {\em \mbf{KA:} Your analysis of the second case is incorrect. 
Can you tell me where the error is in your reasoning: ``If p5 is in seat 2: 
p1 would have to be seated in seat 1 to flank p5. However, there is no available 
seat for p3 to also flank p5 on the other side, as p3 would be required to sit 
in seat 3, which is not adjacent to p5. Thus, this position for p5 is not possible''?
\\[0.05in] \mbf{GPT-4:} 
You are right, and I apologize for the error in my previous analysis. Let's re-examine the second case:
\bit
\item If p5 is in seat 2: p1 would have to be seated in seat 1 to flank p5. p3 could be seated in seat 3, which is also flanking p5.
\eit
In this case, p1 and p3 are indeed flanking p5. My previous analysis contained an error, 
and I apologize for the confusion. Considering this, p5 doesn't necessarily have to be in the middle seat.
} \\[0.05in]
\gsp realizes this particular error when its attention is explicitly brought to it, 
but it still fails to properly reassess the situation, fix the other errors, and arrive at 
even one of the two correct conclusions.

\subsection{Temporal Reasoning}
Here we give \gsp a simple temporal-reasoning problem.\footnote{Formally, this problem belongs to a class of
temporal-reasoning problems literally known as STP (``Simple Temporal Problems'') \cite{Planken2008TemporalRP}. 
This class is of limited
expressivity and there exist very efficient algorithms for solving STPs (\egnsp, consistency can be decided
in $O(n\cdot m)$ where $n$ is the number of events described in a given STP and $m$ is the number of constraints
between the events).} \gsp arrives at a false conclusion, making a number of glaring errors along the way. 
\\[0.05in] {\em \mbf{KA}: 
Tom and Nancy commute to work. Nancy's commute takes about 30 to 40 minutes, while Tom's commute takes about 40 to 50 minutes.
Last Friday, Nancy left home between 8:10 and 8:20 AM,  while Tom arrived at work between 8:50 and 9:10 AM.
In addition, Nancy arrived at work after Tom left his place, but no more than 20 minutes after that.
What can we conclude about when Tom and Nancy arrived at work last Friday?
\mbox{}\\[0.1in] \mbf{GPT-4:} 
Let's break this down:

Nancy's commute takes between 30 to 40 minutes. That means if she left
home at the latest possible time (8:20 AM), she would arrive at work
between 8:50 and 9:00 AM.

Tom's commute takes between 40 to 50 minutes. If he arrived at work at
the earliest possible time (8:50 AM), then the latest he could have
left home is 8:10 AM. If he arrived at work at the latest possible
time (9:10 AM), then the earliest he could have left home is 8:20 AM.

Nancy arrived at work after Tom left home, but not more than 20
minutes after he left.

{\bf From these facts, we can deduce the following:}

If Tom left his home at the latest possible time of 8:20 AM and Nancy
left at her latest possible time of 8:20 AM, and her commute is a
maximum of 40 minutes, Nancy could have arrived at work as late as
9:00 AM. Since Tom's commute is a minimum of 40 minutes, this means
that he would have arrived at work by 9:00 AM at the latest as well.

Conversely, if Tom left home at his earliest possible time (8:10 AM)
and Nancy left at her latest possible time (8:20 AM), even if Nancy's
commute took only 30 minutes, she would arrive at work by 8:50 AM at
the earliest, which is not more than 20 minutes after Tom left. Since
Tom's commute is at least 40 minutes, he would arrive at work by 8:50
AM at the earliest as well.

In conclusion, both Tom and Nancy arrived at work between 8:50 and
9:00 AM last Friday. We can't say for sure who arrived first since
there's overlap in the time ranges, but we do know they both arrived
within that 10-minute window.}
\\[0.05in]
Once again we see that \gsp summarizes/organizes the given information impeccably, 
but fails miserably at using that information to deduce sound and useful conclusions. 
The errors start immediately after the summary is completed and \gsp writes 
``From these facts, we can deduce the following'' (I have boldfaced this sentence in the transcript). It starts by positing 
the following: ``If Tom left his home at the latest possible time of 8:20 AM $\cdots$''
But in fact there are no given constraints about the latest possible time that
Tom left his home; \gsp is taking the constraint that was given about 
Nancy (``Nancy left home between 8:10 and 8:20 AM'') and hallucinating its application
to Tom. Moreover, the entire
conditional statement is bizarre, because its antecedent includes information
about Tom that is completely irrelevant to the conclusion (Nancy's arrival):
``If Tom left his home at the latest possible time of 8:20 AM and Nancy left at her latest possible time of 8:20
AM, and her commute is a maximum of 40 minutes, Nancy could have arrived at work as late as 9:00 AM.''
This should have been stated as: ``If Nancy left at her latest possible time of 8:20
AM, and her commute is a maximum of 40 minutes, Nancy could have arrived at work as late as 9:00 AM.''

\gsp then proceeds to incorrectly infer the following: ``Since Tom's commute is 
a minimum of 40 minutes, this means that he would have arrived at work 
by 9:00 AM at the latest as well.'' This does not follow at all, and it certainly doesn't follow
from the fact that ``Tom's commute is a minimum of 40 minutes.''

The following paragraph is predicated on the hallucinated constraint that Tom's earliest departure
time was 8:10 AM (again, this constraint was given for Nancy, not for Tom). It then goes on to claim 
that 8:50 AM is ``not more than 20 minutes after'' 8:10 AM. Finally, it concludes (wrongly) that
they both arrived between 8:50 and 9:00. 

Remarkably, when a countermodel is explicitly given to it, 
\gsp rejects it (again incorrectly):
{\em  \mbox{}\\[0.05in] \mbf{KA}:  Why is it not possible for Tom to leave at 8:30 and arrive at
9:10, and for Nancy to leave at 8:15 and arrive at 8:45?  
\mbox{}\\[0.05in] \mbf{GPT-4:} 
I apologize for any misunderstanding. From the information you initially
provided, it was stated that Nancy's commute takes between 30 to 40
minutes and Tom's commute takes between 40 to 50 minutes. 

If Tom were to leave at 8:30 AM, with a minimum commute time of 40
minutes, the earliest he could arrive at work would be 9:10 AM. So,
Tom leaving at 8:30 AM and arriving at 9:10 AM is a possibility within
the given constraints.

However, if Nancy left at 8:15 AM, even with her shortest commute time
of 30 minutes, the earliest she could arrive at work would be 8:45
AM. This would mean that she would arrive at work exactly 30 minutes
after Tom left home. But according to the conditions provided, Nancy
should arrive at work not more than 20 minutes after Tom leaves his
home. Therefore, the scenario you suggested with Nancy leaving at 8:15
and arriving at 8:45 is not possible within the provided constraints.

I hope this clears up the confusion. If you have further questions or
if there is more information to consider, please let me know!
}  \\[0.05in]
\gsp is sticking to its guns because apparently it is convinced that 8:45 
is ``exactly 30 minutes'' after 8:30. 
\\[0.05in] {\em \mbf{KA}:  But 8:45 (Nancy's arrival) is only 15 minutes after 
8:30, which is when Tom left his place, not 30 minutes as you claim.  
\mbox{}\\[0.05in] \mbf{GPT-4:} I apologize for the oversight. You're correct.

If Nancy left at 8:15 AM and her commute took 30 minutes, she would
indeed arrive at 8:45 AM, which is 15 minutes after Tom left at 8:30
AM (if we assume this scenario). This fits within the constraint that
Nancy must arrive at work not more than 20 minutes after Tom leaves
his home. And with a 40-minute commute, Tom would indeed arrive at
9:10 AM.

So, you're correct. It is indeed possible for Tom to leave at 8:30 AM
and arrive at 9:10 AM, and for Nancy to leave at 8:15 AM and arrive at
8:45 AM. Thank you for pointing out the error. I appreciate your
patience and apologize for any confusion caused.
}

\subsection{Murder or Suicide?}
This is a logic puzzle I made up a while back. The conclusion is that Aunt Agatha killed herself. This follows by eliminating Charles and the butler. First, Aunt Agatha must have hated herself, because she hated everyone other than the butler. Therefore, Charles did not hate her (since he doesn't hate anyone that Aunt Agatha hates), and hence he could not have killed her (by premise 3). The butler could not hate himself, because if he did, he would hate everyone (since he already hates everyone else, through premises 5 and 7), and we know that's not possible by premise 8. Thus, the butler must be richer than Aunt Agatha, or else he would hate himself (by premise 6), which means he could not be the killer (premise 3). 

\gsp correctly rules out Charles, but is unable to rule out the butler and comes to incorrectly deduce that he is the killer.
Another key---and rather strange---mistake that \gsp makes is this: ``Since Aunt Agatha hates everyone except the butler (premise 5), this
means she must at least not hate herself.'' This is an odd error, because the natural conclusion to draw from premise 5 is that
Aunt Agatha hates herself. Along the way, \gsp displays the internal inconsistency we've seen repeatedly in the other examples. 
Virtually in every single response, \gsp claims to derive some proposition as well as its negation. 
\\[0.05in] {\em \mbf{KA}: You are given the following premises:
\ben
\item Someone who lives in Dreadbury Mansion killed Aunt Agatha.
\item The only people who live in Dreadbury Mansion are Aunt Agatha, the butler,
and Charles.
\item A killer always hates his victims, and is never richer than his victims.
\item Charles hates no one that Aunt Agatha hates.
\item Aunt Agatha hates everyone except the butler.
\item The butler hates everyone not richer than Aunt Agatha.
\item The butler hates everyone Aunt Agatha hates.
\item No one hates everyone.
\item Aunt Agatha is not the butler.
\een
On the basis of this information, determine who killed Aunt Agatha and give a
detailed proof that your conclusion follows from the premises.
\mbox{}\\[0.1in] \mbf{GPT-4:} Let's analyze each of the premises:

Someone who lives in Dreadbury Mansion killed Aunt Agatha.  The only
people who live in Dreadbury Mansion are Aunt Agatha, the butler, and
Charles.  From these two premises, we can deduce that Aunt Agatha was
either killed by Charles or the butler, because they are the only
other residents of the Dreadbury Mansion.

A killer always hates his victims, and is never richer than his
victims.  This premise is crucial as it gives us two conditions for
someone to be a killer: they must hate the victim and not be richer
than them.

Charles hates no one that Aunt Agatha hates.  Aunt Agatha hates
everyone except the butler.  From premises 4 and 5, we can infer that
Charles does not hate Aunt Agatha because Aunt Agatha hates herself
(since she hates everyone except the butler) and Charles hates no one
that Aunt Agatha hates.

So, by the rules of premise 3, Charles cannot be the killer because he
does not hate Aunt Agatha.

The butler hates everyone not richer than Aunt Agatha.  The butler
hates everyone Aunt Agatha hates.  Premise 6 indicates that the butler
hates Aunt Agatha, as she is not richer than herself. And from premise
7, we understand that the butler also hates Aunt Agatha because Aunt
Agatha hates herself (as she hates everyone except the butler). So,
the butler does indeed hate Aunt Agatha.

However, this doesn't make him the killer yet. We still need to verify
the second condition from premise 3, that the killer is not richer
than the victim. As there is no information regarding their wealth, we
must assume that the butler could either be richer or poorer than Aunt
Agatha.

No one hates everyone.  Aunt Agatha is not the butler.  Premise 9 is
not immediately useful, but premise 8 tells us that no one can hate
everyone. Since Aunt Agatha hates everyone except the butler (premise
5), this means she must at least not hate herself. This confirms the
idea that Aunt Agatha did not commit suicide.

From this analysis, Charles can't be the killer because he doesn't
hate Aunt Agatha. Aunt Agatha can't be her own killer because she
doesn't hate herself (and thus didn't commit suicide). Therefore, the
only remaining person who can be the killer is the butler.

Thus, the killer of Aunt Agatha must be the butler.
\\[0.05in] \mbf{KA}: You've concluded both that Aunt Agatha hates herself (you wrote
``Aunt Agatha hates herself (since she hates everyone except the
butler)'') and also that she doesn't hate herself (``Since Aunt Agatha
hates everyone except the butler (premise 5), this means she must at
least not hate herself''). Can both of these be true? Or are the
premises inconsistent?
\\[0.05in] \mbf{GPT-4}: You're correct. There seems to be a
mistake in my reasoning. It's contradictory to state both that Aunt
Agatha hates herself and doesn't hate herself. Let's try to clear up
this inconsistency:

From premise 5 (``Aunt Agatha hates everyone except the butler.''), we
infer that Aunt Agatha doesn't hate herself as ``everyone'' typically
refers to every other person, not including oneself.

Using this corrected interpretation, let's revisit premise 7 (``The
butler hates everyone Aunt Agatha hates.''). Since Aunt Agatha doesn't
hate herself, the butler doesn't hate Aunt Agatha. Thus, according to
premise 3 (``A killer always hates his victims, and is never richer
than his victims.''), the butler cannot be the killer.

So we are left with Charles. Based on premise 4 (``Charles hates no one
that Aunt Agatha hates.''), we know that Charles also doesn't hate Aunt
Agatha, because Aunt Agatha doesn't hate herself. Therefore, Charles
cannot be the killer, as per premise 3.

We thus have a contradiction. According to the premises, neither
Charles nor the butler can be the killer, but Aunt Agatha must have
been killed by someone who lives in Dreadbury Mansion (premise 1 and
2). It appears the premises as given are inconsistent.
\\[0.05in] \mbf{KA}: First, ``everyone'' here means literally everyone, in the sense of
formal logic (there is a universal quantifier involved). Second, you
said earlier that the butler was the killer. Why are you now ruling out
the butler?
\\[0.05in] \mbf{GPT-4}: I apologize for the confusion. If we are
interpreting ``everyone'' as literally everyone, including oneself, then
Aunt Agatha hates herself. This is a different understanding from my
earlier analysis.

Let's go through the premises again under this new understanding:

Aunt Agatha, the butler, and Charles live in Dreadbury
Mansion. (Premise 2) The killer must be one of these three because
someone who lives in Dreadbury Mansion killed Aunt Agatha. (Premise 1)
Aunt Agatha hates everyone including herself. (Premise 5, under our
new interpretation of ``everyone'') The butler hates everyone that Aunt
Agatha hates, so the butler hates Aunt Agatha. (Premise 7) Charles
hates no one that Aunt Agatha hates, so Charles does not hate Aunt
Agatha. (Premise 4) From these points, according to Premise 3 (``A
killer always hates his victims, and is never richer than his
victims''), we can already eliminate Charles as the killer since he
does not hate Aunt Agatha.

So, the killer must be either Aunt Agatha or the butler.

But since no one hates everyone (Premise 8), we can infer that Aunt
Agatha doesn't hate herself, and thus she wouldn't have killed
herself. So, Aunt Agatha is not the killer.

That leaves us with the butler as the only possible killer. So, even
with our new understanding of ``everyone'' as literally everyone, the
conclusion remains: The butler killed Aunt Agatha. I apologize for the
previous error in reasoning, and thank you for your patience.
\\[0.05in] \mbf{KA}: You're again deriving two inconsistent conclusions: that Agatha
hates herself and that she doesn't. Which is it?
\\[0.05in] \mbf{GPT-4}: You're right, my apologies for the confusion. The premises
do seem to be leading to inconsistent conclusions regarding whether
Aunt Agatha hates herself or not.

Let's go through this more carefully.

Premise 5: ``Aunt Agatha hates everyone except the butler.''  Given your
interpretation, ``everyone'' here includes herself, which means Aunt
Agatha hates herself.

However, Premise 8 states: ``No one hates everyone.''  Given your
interpretation of ``everyone'' as literally everyone, then Aunt Agatha
does not hate herself. She must have at least one exception aside from
the butler, or else she would hate ``everyone,'' which contradicts
Premise 8.

Therefore, Aunt Agatha does not hate herself. The only interpretation
of the premises that does not lead to a contradiction is that Aunt
Agatha does not hate herself.

Now, let's reconsider who the killer can be:

Aunt Agatha did not kill herself (because she does not hate herself).
Charles did not kill Aunt Agatha (because he does not hate anyone that
Aunt Agatha hates, which includes herself).  The butler did not kill
Aunt Agatha (because he does not hate anyone that Aunt Agatha does not
hate, and Aunt Agatha does not hate herself).  Therefore, under these
premises and your interpretation of ``everyone,'' none of Aunt Agatha,
the butler, or Charles could have killed Aunt Agatha.

In conclusion, there is a contradiction: Premises 1 and 2 state that
Aunt Agatha must have been killed by someone in Dreadbury Mansion,
which includes only Aunt Agatha, the butler, and Charles. But, under
our analysis, none of them could have killed Aunt Agatha. This
suggests the premises are inconsistent.
\\[0.05in] \mbf{KA}: Let's focus on this piece of your analysis:

'However, Premise 8 states: ``No one hates everyone.''  Given your
interpretation of ``everyone'' as literally everyone, then Aunt Agatha
does not hate herself. She must have at least one exception aside from
the butler, or else she would hate ``everyone,'' which contradicts
Premise 8.'

Can you prove in more detail, step-by-step, how premise 8 entails that
Aunt Agatha does not hate herself?
\\[0.05in] \mbf{GPT-4}: Sure, I'll clarify this deduction in more detail:

Premise 8 states: ``No one hates everyone.'' In the context of formal
logic, ``everyone'' typically refers to all individuals within a certain
set. Given your interpretation that ``everyone'' should be taken
literally as all individuals without exception, ``everyone'' includes
each individual person, and themselves.

Therefore, according to Premise 8, there is no individual who hates
every other individual and also hates themselves.

Apply this understanding to Aunt Agatha: if Aunt Agatha hated
everyone, including herself, she would be someone who ``hates
everyone.'' But this would contradict Premise 8, which states that no
such person exists.

Therefore, Aunt Agatha does not hate everyone.

In order to not hate ``everyone'', there must be at least one individual
whom Aunt Agatha does not hate. We know from Premise 5 that the butler
is one such individual.

However, even if Aunt Agatha does not hate the butler, she would still
be ``hating everyone'' if she hated all other individuals, including
herself. This again would contradict Premise 8.

Therefore, there must be at least one more individual, in addition to
the butler, whom Aunt Agatha does not hate. Since the only other
individuals in question are Charles and Aunt Agatha herself, and we
know from Premise 5 that she hates Charles, the only remaining
possibility is that Aunt Agatha does not hate herself.

Therefore, Premise 8 implies that Aunt Agatha does not hate herself.
\\[0.05in] \mbf{KA}: Justify step 6 please: 'However, even if Aunt Agatha does not
hate the butler, she would still be ``hating everyone'' if she hated all
other individuals, including herself.''
Why does that hold?
\\[0.05in] \mbf{GPT-4}: Sure, let's clarify step 6:

According to your interpretation of ``everyone,'' this term refers to
all individuals, including oneself. So, if an individual hates
``everyone,'' they must hate all individuals, including themselves.

So in step 6, when we say ``if Aunt Agatha hated all other individuals,
including herself, she would still be 'hating everyone','' what we're
saying is:

If Aunt Agatha hated herself and all other individuals (except for the
butler), then she would hate all individuals in our context (which
includes Aunt Agatha, the butler, and Charles), with the sole
exception of the butler.

However, even with this exception, she would still be classified as
someone who hates ``everyone'' under our interpretation, because she
hates all individuals including herself (despite the exception of the
butler).

Thus, even if Aunt Agatha does not hate the butler, if she hated all
other individuals, including herself, she would be classified as
someone who ``hates everyone'' (despite the one exception). This would
contradict Premise 8, which states that no one hates everyone.

Therefore, if Aunt Agatha does not hate the butler (as per Premise 5)
and she is to avoid contradicting Premise 8, there must be at least
one more individual whom Aunt Agatha does not hate. Since the only
other individuals in our context are Charles and Aunt Agatha herself,
and we know from Premise 5 that she hates Charles, the only remaining
possibility is that Aunt Agatha does not hate herself.

I hope that clarifies step 6. It's a nuanced point: even if Aunt
Agatha has one exception to her hatred (the butler), if she hates all
other individuals, including herself, she would still be considered as
someone who ``hates everyone'' under our interpretation, which
contradicts Premise 8.
\\[0.05in] \mbf{KA}: This is quite wrong. As long as there is at least one individual
b not hated by Aunt Agatha, then she cannot possibly hate everyone,
period, because if she did hate everyone then she would also hate b,
by definition of 'everyone'; and we know that she does not hate b. In
our case, of course, the individual b in question is the butler.
\\[0.05in] \mbf{GPT-4}: You're absolutely right, and I apologize for my
misinterpretation. I appreciate your patience.

Given your correct interpretation, if Aunt Agatha does not hate the
butler, it is sufficient to say that she doesn't hate everyone, even
if she hates all others including herself.

This also affects the conclusions we draw from other premises.

Let's look at premises 3, 4, and 7 again:

Premise 3 states: ``A killer always hates his victims, and is never
richer than his victims.''  Premise 4 states: ``Charles hates no one
that Aunt Agatha hates.'' Premise 7 states: ``The butler hates everyone
Aunt Agatha hates.'' From premise 4, since Charles hates no one that
Aunt Agatha hates, and Aunt Agatha hates herself, it follows that
Charles does not hate Aunt Agatha. This means that Charles could not
have killed Aunt Agatha (as per premise 3).

From premise 7, since the butler hates everyone that Aunt Agatha
hates, and Aunt Agatha hates herself, it follows that the butler hates
Aunt Agatha. Hence, he could be the killer, according to the hatred
condition of premise 3.

For the wealth condition of premise 3, there is no clear
information. We must assume that Aunt Agatha was not richer than
everyone else in the mansion; if she was, the premise ``a killer is
never richer than his victims'' would make it impossible for anyone to
have killed her. This is a reasonable assumption given that the other
premises suggest there is a killer.

So, we have only one viable candidate: the butler. This aligns with
our conclusion from the first analysis: The butler is the one who
killed Aunt Agatha.

Thank you for your guidance in clarifying these premises.
\\[0.05in] \mbf{KA}: Why did you say earlier that even if there is an individual b not
hated by Aunt Agatha, it would still follow that Aunt Agatha hates
everyone?
\\[0.05in] \mbf{GPT-4}: You're correct, and I must apologize for the
confusion. That statement was a misinterpretation on my part. In
formal logic, if there is an individual that Aunt Agatha does not hate
(in this case, the butler), it does not follow that Aunt Agatha hates
everyone. In fact, it's the opposite: if there is at least one
individual Aunt Agatha does not hate, it means she does not hate
everyone. I appreciate your patience and I'm sorry for the oversight.
\\[0.05in] \mbf{KA}: Ok, so you are now convinced that the butler killed Aunt Agatha?
\\[0.05in] \mbf{GPT-4}: Given the premises and the interpretation of ``everyone'' as
literally everyone, including oneself, the evidence does indeed point
to the butler as the killer. Here's why:

According to premise 5, Aunt Agatha hates everyone except the
butler. Given the interpretation of ``everyone'', this means she hates
herself.

According to premise 7, the butler hates everyone Aunt Agatha hates,
which now includes Aunt Agatha herself.

Charles, according to premise 4, hates no one that Aunt Agatha hates,
so he does not hate Aunt Agatha.

From premise 3, a killer always hates his victims. So, Charles, who
does not hate Aunt Agatha, can't be the killer. The butler, who does
hate Aunt Agatha, could be the killer. And, as Aunt Agatha is not the
butler (premise 9), and she can't have killed herself since a killer
hates his victims and Aunt Agatha doesn't hate herself (from premise
8), Aunt Agatha herself is not the killer.

So, we can conclude that the butler is the killer, given that he is
the only person who fits the description provided by the
premises. This conclusion depends heavily on the interpretation of
``everyone'' to include oneself, which allows us to infer that Aunt
Agatha hates herself.
\\[0.05in] \mbf{KA}: You said earlier that there's not enough information in the
premises about the butler's level of wealth. Are you sure there's
nothing we can infer?
\\[0.05in] \mbf{GPT-4}: You're correct, and I apologize for
missing that detail. There is indeed something we can infer about the
butler's wealth from the given premises.

Premise 3 states that ``A killer always hates his victims, and is never
richer than his victims.''

We know from premises 5 and 7 that Aunt Agatha hates herself and,
thus, the butler also hates Aunt Agatha. If the butler was richer than
Aunt Agatha, according to premise 3, he could not be her killer.

However, premise 6 states that ``The butler hates everyone not richer
than Aunt Agatha.'' Given that we know the butler hates Aunt Agatha, it
follows that Aunt Agatha must not be richer than the butler, or else
he wouldn't hate her (which would contradict premises 5 and 7).

In summary, the butler isn't richer than Aunt Agatha, but Aunt Agatha
isn't richer than the butler either. This suggests they are of equal
wealth.

So, given all of the premises, the butler, who hates Aunt Agatha and
is of equal wealth, is indeed the only individual who fits the profile
of the killer.

Thank you for your patience in clarifying these premises. I appreciate
your guidance in this logical puzzle.
\\[0.05in] \mbf{KA}: So what did you conclude about the butler's wealth, that he
neither richer nor poorer than Aunt Agatha?
\\[0.05in] \mbf{GPT-4}: 
Yes, from the
given premises, we can infer that the butler is neither richer nor
poorer than Aunt Agatha, which would suggest they are of equal
wealth. Here's how:

Premise 6 states, ``The butler hates everyone not richer than Aunt
Agatha.'' This means that if someone is as rich as Aunt Agatha or
poorer, the butler hates them.

We know from premises 5 and 7 that the butler hates Aunt
Agatha. Therefore, according to premise 6, Aunt Agatha must not be
richer than the butler. If she were, the butler wouldn't hate her,
contradicting premises 5 and 7.

At the same time, the butler can't be richer than Aunt Agatha
either. That's because premise 3 states that ``a killer is never richer
than his victims.'' If the butler were richer than Aunt Agatha, he
couldn't be her killer.

So, the butler isn't richer than Aunt Agatha, and Aunt Agatha isn't
richer than the butler. This suggests that they are of equal
wealth. Therefore, given all the premises, the butler fits the
criteria to be the killer: he hates Aunt Agatha (as required by
premise 3), and he isn't richer than her (also required by premise 3).
\\[0.05in] \mbf{KA}: I claim that we can infer that the butler is richer than Aunt
Agatha.
\\[0.05in] \mbf{GPT-4}: You're correct, and I apologize for the oversight.

Premise 6 states that ``The butler hates everyone not richer than Aunt
Agatha.'' We know from premises 5 and 7 that the butler hates Aunt
Agatha. Therefore, it must be the case that Aunt Agatha is not richer
than the butler, i.e., the butler is either as rich as or richer than
Aunt Agatha.

However, premise 3 states that a killer is never richer than his
victims. If the butler was as rich as Aunt Agatha, he could still be
her killer. But since Aunt Agatha was killed by someone living in
Dreadbury Mansion (premise 1) and Charles and Aunt Agatha herself
could not have been the killers (as they don't hate Aunt Agatha), the
butler must be the killer. Given that the butler is the killer and
that a killer is never richer than his victims, it must be the case
that the butler is not richer than Aunt Agatha.

The only way to satisfy both conditions (the butler being neither
richer than nor as rich as Aunt Agatha) is if the butler is poorer
than Aunt Agatha.

I apologize for the confusion in the previous responses. The
conclusion should be that the butler, who is poorer than Aunt Agatha,
killed her. This conclusion is consistent with all the given premises.

Again, I appreciate your patience in solving this logical puzzle.
\\[0.05in] \mbf{KA}: No, like I said, it's the opposite. We can infer that the butler,
far from being poorer than Aunt Agatha, he's actually richer.
\\[0.05in] \mbf{GPT-4}: You're correct. I apologize for the confusion. Let's revisit premise
6: ``The butler hates everyone not richer than Aunt Agatha.''

This means that the butler hates all individuals who are as wealthy as
Aunt Agatha or less wealthy than her. But from premises 5 and 7, we
know that the butler does indeed hate Aunt Agatha. Therefore, Aunt
Agatha must be in the group of people who are not richer than the
butler, which implies that the butler is as wealthy as or wealthier
than Aunt Agatha.

Given premise 3, which states that a killer is never richer than his
victims, it would initially seem that the butler cannot be the killer
since he could potentially be richer than Aunt Agatha. However, Aunt
Agatha cannot be the killer because she doesn't hate herself (from
premise 8), and Charles can't be the killer because he doesn't hate
Aunt Agatha (from premises 4 and 5).

Hence, the butler must be the killer, and the only way the butler can
be the killer and still satisfy premise 3 is if the butler and Aunt
Agatha have the same wealth. Thus, there was a mistake in my previous
analysis: the butler isn't poorer than Aunt Agatha. Instead, the
butler must be equally wealthy as Aunt Agatha. However, your claim was
that we can infer the butler is richer than Aunt Agatha, which is not
supported by the premises.

I apologize for the confusion, and I appreciate your patience and
assistance in resolving this puzzle.
\\[0.05in] \mbf{KA}: So you're back to claiming that the butler is equally rich as
Aunt Agatha, and that my claim that he's richer is does not follow.
\\[0.05in] \mbf{GPT-4}: My apologies for the confusion in my previous responses. Let's
reevaluate premise 6: ``The butler hates everyone not richer than Aunt
Agatha.''

From this premise, we can infer that the only people whom the butler
doesn't hate are those richer than Aunt Agatha. But since we know from
premises 5 and 7 that the butler does hate Aunt Agatha, it must follow
that Aunt Agatha is not richer than the butler. Therefore, the butler
must be richer than Aunt Agatha.

In light of this, we return to premise 3, which states, ``A killer
always hates his victims, and is never richer than his victims.'' Given
that the butler hates Aunt Agatha and is richer than her, it would
seem at first that the butler cannot be the killer.

However, if we consider all the residents of Dreadbury Mansion, Aunt
Agatha is not the killer because she doesn't hate herself (from
premise 8) and Charles is not the killer because he doesn't hate Aunt
Agatha (from premises 4 and 5).

Hence, we're left with the butler as the only possible killer, despite
him being richer than Aunt Agatha, which seems to contradict premise
3. This seems to create a paradox given the premises as they stand.

There may be an issue with how we're interpreting the premises or
there may be an error in the premises themselves. If the butler is
indeed the killer, then it contradicts the wealth condition stated in
premise 3.

If we adhere strictly to the premises, it appears we have a
paradox. The butler seems to be the only possible candidate who could
be the killer based on hatred, but this contradicts the wealth
condition of the killer in premise 3. I appreciate your patience and I
apologize for the confusion in my previous responses.
\\[0.05in] \mbf{KA}: No, this reasoning of yours is incorrect: ``But since we know
from premises 5 and 7 that the butler does hate Aunt Agatha, it must
follow that Aunt Agatha is not richer than the butler. Therefore, the
butler must be richer than Aunt Agatha.''

The fact that Aunt Agatha is not richer than the butler doesn't
support the conclusion that the butler is richer. I still claim that
the butler IS in fact richer, but your attempts to prove that fact
have all contained reasoning errors.
\\[0.05in] \mbf{GPT-4}:
You're correct, and I
apologize for the confusion. Your original claim was that the butler
is richer than Aunt Agatha, not equally rich as her or poorer. Let's
revisit the premises to see how we can support this claim:

From Premise 6 (``The butler hates everyone not richer than Aunt
Agatha''), we can infer that the butler does not hate people who are
richer than Aunt Agatha.  From Premise 7 (``The butler hates everyone
Aunt Agatha hates''), coupled with the information from Premise 5
(``Aunt Agatha hates everyone except the butler''), we know the butler
hates Aunt Agatha.  Combining these pieces of information, we can
conclude that Aunt Agatha is not richer than the butler because if she
were, the butler wouldn't hate her (which would contradict Premise 7).
However, this doesn't automatically mean that the butler is richer
than Aunt Agatha. It could also mean they have the same level of
wealth.  But if we look at Premise 3 (``A killer always hates his
victims, and is never richer than his victims''), we know that the
killer is not richer than Aunt Agatha.  Given all the premises, we've
deduced that the killer can't be Aunt Agatha herself or Charles,
leaving only the butler as a possible killer. If the butler were the
killer, and the killer can't be richer than Aunt Agatha, then the
butler can't be richer than Aunt Agatha.  Hence, we have a
contradiction, since we earlier deduced from Premise 6 that Aunt
Agatha can't be richer than the butler. The only resolution is that
the butler must be richer than Aunt Agatha.  In conclusion, I
apologize for the confusion in the earlier responses. Your claim is
correct: from the given premises, it can be inferred that the butler,
who is richer than Aunt Agatha, is the one who killed her.
\\[0.05in] \mbf{KA}: That's impossible, because a killer is never richer than his
victims.
\\[0.05in] \mbf{GPT-4}: I apologize for the confusion in my previous
messages. Let's go through the premises again. \\
$[\cdots \; \mbox{Similar confusions/errors as above\/}\; \cdots]$
}

\subsection{Wason Selection Task}
The Wason selection task is a staple in the psychology of reasoning. 
The January paper included an example that GPT-3.5 flunked. Here is 
another version that GPT-4.0 fails badly: 
\\[0.05in] {\em \mbf{KA}: 
Seven cards are placed on the table, each of which has a number on one side and a single colored patch on the other side.
The faces of the cards show 50, 16, red, yellow, 23, green, 30. Which cards would you have to turn to test the truth
of the proposition that if a card is showing a multiple of 4 then the color of the opposite side is yellow? 
\mbox{}\\[0.1in] \mbf{GPT-4:} 
The proposition is: ``If a card shows a number that is a multiple of 4,
then the color on the other side is yellow''. In order to test this
proposition, you would have to turn over the following cards:

Card ``16'': This card is showing a number that is a multiple of
4. According to the proposition, the color on the other side should be
yellow. By turning over this card, you can check whether this part of
the proposition holds true.

Card ``Yellow'': This card is showing a color that according to the
proposition should be on the other side of a card with a number that
is a multiple of 4. By turning it over, you can check if the number on
the other side is a multiple of 4.

Card ``50'' and ``30'': While these cards show numbers that are not
multiples of 4, they should be checked to ensure the color on the
other side is not yellow. If the other side is yellow, then it would
disprove the proposition that only multiples of 4 can have yellow on
the other side.

You wouldn't need to turn over the card ``23'', ``red'', or ``green''. The
card ``23'' is not a multiple of 4, so it doesn't matter what color is
on the other side based on the given proposition. The ``red'' and
``green'' cards, if they have numbers that are multiples of 4 on the
other side, would not disprove the proposition because the proposition
does not state that yellow is the only color that can be on the other
side of a card with a number that's a multiple of 4.
} \\[0.05in]
In fact, only 16, red, and green need to be turned over. Thus, in terms of precision, 
\gsp gets 1/4 = 0.25, while on recall it gets 1/3, for a combined
F-score of 0.28. The responses show yet again that \gsp does not understand 
the semantics of material conditionals. When \gsp says that cards ``50'' and ``30''
must be turned over, it seems to mistake the \temv{if} for an {\em if and only if}: 
``{\em while these cards show numbers that are not multiples of 4, they should be 
checked to ensure the color on the other side is not yellow. If the other side is yellow, 
then it would disprove the proposition that only multiples of 4 can have yellow on
the other side}''. But of course the proposition being tested does not require that
\temv{only} multiples of 4 have yellow on the other side; it simply requires that 
{\em if\/} a card has a multiple of 4 on the other side, \temv{then} the other side
must be yellow; or, equivalently, that if a card face is \temv{not} yellow then the
number on the other side is not a multiple of 4. 

Observe, however, that even the mistaken interpretation of the rule as a biconditional
is {\em not consistently observed}. If \gsp understood the rule being tested as a biconditional
and hence felt compelled to turn over 50 and 30, {\em then it ought to have done the same thing
for 23}. But it didn't. In fact, it explicitly said that {\em The card ``23'' is not a 
multiple of 4, so it doesn't matter what color is on the other side based on the given 
proposition}, which contradicts the biconditional interpretation that it used as a rationale
for turning 50 and 30. This is one more illustration of the other major theme emerging
from these examples: The responses of \gnsp,  whether right or wrong, are often internally inconsistent. 
Yet another example appears in the following problem.

\subsection{Entropy}
An elementary result of information theory states that the entropy of a random vector $Z$ is bounded above by the sum of the entropies
of the random variables that comprise $Z$. Hence, the answer to the following question should be ``under no conditions'': \\[0.05in]
{\em \mbf{KA}: Let Z be a random vector consisting of n random variables $X_1$,...$X_n$.  
Under what conditions can the entropy of Z exceed the sum of the entropies of all $X_i$?
\mbox{}\\[0.1in] \mbf{GPT-4:} 
The entropy of a random vector Z consisting of n random variables $X_1$, ..., $X_n$ 
can exceed the sum of the entropies of all $X_i$ only when there is some dependency 
between the random variables. If the random variables are mutually independent, 
the entropy of Z would be equal to the sum of the entropies of all $X_i$.
The entropy of a random vector Z, H(Z), can be written as the joint entropy of the 
random variables $X_1, ..., X_n$: $H(Z) = H(X_1, ..., X_n)$.
The joint entropy $H(X_1, ..., X_n)$ can be greater than the sum of individual 
entropies $H(X_i)$ 
if and only if the random variables are not mutually independent. In other words, 
knowing the value of one or more of the random variables provides information about 
the others. If the random variables are mutually independent, then the joint entropy 
is equal to the sum of the individual entropies:
\[H(X_1, ..., X_n) = H(X_1) + ... + H(X_n).\]
When there is dependence between the random variables, we can write:
$H(X_1, ..., X_n) \leq H(X_1) + ... + H(X_n)$
with equality if and only if the random variables are mutually independent.
To summarize, the entropy of the random vector Z can exceed the sum of the entropies of
all $X_i$ if and only if there is some dependency between the random variables $X_1, ..., X_n$.
} \\[0.05in]
Observe the internal inconsistency again: It's clear that \gsp has memorized the result correctly:
\[H(X_1, \ldots, X_n) \leq H(X_1) + \cdots + H(X_n),\]
with equality holding iff $X_1,\ldots,X_n$ are mutually independent. Nevertheless, it does not have the
reasoning ability to realize what this entails, namely, that it precludes: 
\[H(X_1, \ldots, X_n) > H(X_1) + \cdots + H(X_n).\]

At least here once the error is explicitly pointed out to \gsp, 
it acknowledges it and corrects itself: 
\\[0.05in] {\em \mbf{KA}: But if $H(Z) = H(X_1, ..., X_n)$ and $H(X_1, ..., X_n) \leq H(X_1) + ... + H(X_n)$ 
then how can we ever have $H(Z) > H(X_1) + ... + H(X_n)$?
\mbox{}\\[0.1in] \mbf{GPT-4:} 
My apologies, I made an error in my previous response. 
You are right, the entropy of the random vector Z cannot exceed the sum of the entropies of all $X_i$. 
The correct statement should be:
When the random variables are mutually independent, the joint entropy H(Z) is equal to the sum 
of the individual entropies: $H(Z) = H(X_1) + ... + H(X_n)$ When there is dependence between the 
random variables, the joint entropy H(Z) is less than the sum of the individual entropies:
$H(Z) < H(X_1) + ... + H(X_n)$
I apologize for the confusion, and I appreciate your attention to detail.
}

\subsection{Simple Compiler Correctness}
The last reasoning problem we give to \gsp is the most challenging one: It is to prove the correctness of a simple expression compiler. 
Remarkably, \gsp goes about this in the right sort of way, by setting up a structural induction over the abstract grammar of expressions. 
This is no doubt because it has seen similar proofs before, as this is a common type of exercise in 
courses and textbooks on programming language theory.\footnote{This particular version is taken 
from Chapter 18 of the textbook {\em Fundamental Proof Methods in Computer Science\/} by \cite{FPMICS}.}
However, even though the proof is on the right general track, it has several errors. 
(For the record, the compiler is indeed correct, although proving this requires strengthening the induction hypothesis). 
\\[0.05in] {\em \mbf{KA}: 
Suppose I have an abstract grammar for numeric expressions defined as follows: 
\begin{verbatim}
Exp := const(Int) 
     | sum(Exp,Exp) 
     | diff(Exp,Exp) 
     | mult(Exp,Exp) 
     | quot(Exp,Exp)
\end{verbatim}
where {\tt Int} denotes the domain of integers. The semantics of these expressions are defined 
via this interpreter function:
\begin{verbatim}
I: Exp -> Int

I(const(n)) = n
I(sum(e1,e2)) = I(e1) + I(e2)
I(diff(e1,e2)) = I(e1) - I(e2)
I(prod(e1,e2)) = I(e1) * I(e2)
I(quot(e1,e2)) = I(e1) / I(e2)
\end{verbatim}
I now define a virtual machine that executes simple programs that are sequences of commands, where commands
have the following structure:
\begin{verbatim}
Cmd := push(Int) | add | sub | mult | div
\end{verbatim}
The operational semantics of these programs (sequences of commands) are defined by an execution function exec
that takes a program and a stack of integers S and produces an integer as output. Let's
write {\tt []} and {\tt n::S} to indicate the empty stack and the stack obtained by prepending integer {\tt n} to stack {\tt S}, 
respectively. Here's the definition of exec: 
\begin{verbatim}
exec([],n::S) = n
exec(push(n)::C,S) = exec(C,n::S)
exec(add::C,n::m::S) = exec(C,(n+m)::S)
exec(sub::C,n::m::S) = exec(C,(n-m)::S)
exec(mult::C,n::m::S) = exec(C,(n*m)::S)
exec(div::C,n::m::S) = exec(C,(n/m)::S)
\end{verbatim}
Finally, I define a compiler that translates an expression e into a program (sequence of commands)
as follows. I write {\tt @} for sequence concatenation: 
\begin{verbatim}
T: Exp -> List(Cmd)

T(const(n)) = [push(n)]
T(sum(e1,e2)) = T(e2)@T(e1)@[add]
T(diff(e1,e2)) = T(e2)@T(e1)@[sub]
T(prod(e1,e2)) = T(e2)@T(e1)@[mult]
T(quot(e1,e2)) = T(e2)@T(e1)@[div]
\end{verbatim}
Disprove or prove the following claim: For all expressions {\tt e}, {\tt exec(T(e),[]) = I(e)}. 
\\[0.05in] \mbf{GPT-4:} 
The claim is correct. The execution of the translated commands from an expression results 
in the same output as interpreting the expression directly. We can demonstrate this 
using structural induction over the set of expressions.

Inductive hypothesis: For all expressions {\tt e}, {\tt exec(T(e),[]) = I(e)}. 

Base case: Consider the simplest possible expression, a constant integer expression {\tt e = const(n)}.

For {\tt T(const(n))}, we have {\tt T(const(n)) = [push(n)]}. 
Therefore,  
\begin{verbatim}
exec(T(const(n)),[]) = exec([push(n)],[]) 
                     = n (according to the definition of exec).
\end{verbatim}
For {\tt I(const(n))}, we have {\tt I(const(n)) = n}.

So, for the base case, we have shown {\tt exec(T(e),[]) = I(e)}.

Inductive step: We need to consider the other four kinds of expressions: {\tt sum}, {\tt diff}, {\tt prod}, and {\tt quot}. \\
{\tt sum(e1, e2)}: For {\tt T(sum(e1,e2))}, we have {\tt T(sum(e1,e2)) = T(e2)@T(e1)@[add]}. 

Therefore, 
\begin{verbatim}
exec(T(sum(e1,e2)),[])                  =  
exec(T(e1)@T(e2)@[add],[])              = 
exec(T(e2)@T(e1),[]) + exec([add],[])   =
exec(T(e1),[]) + exec(T(e2),[])         (by definition of exec).
\end{verbatim}
For {\tt I(sum(e1,e2))}, we have {\tt I(sum(e1,e2)) = I(e1) + I(e2)}.

By the inductive hypothesis, {\tt exec(T(e1),[]) = I(e1)} and {\tt exec(T(e2),[]) = I(e2)}. 
So 
\begin{verbatim}
exec(T(e1),[]) + exec(T(e2),[]) = I(e1) + I(e2). 
\end{verbatim}
Thus, we have shown that for {\tt sum(e1,e2)}, {\tt exec(T(e),[]) = I(e)}.

{\tt diff(e1, e2)}: A similar reasoning as for sum can be applied to show that for {\tt diff(e1,e2)}, {\tt exec(T(e),[]) = I(e)}.

{\tt prod(e1, e2)}: Likewise, a similar reasoning as for sum can be applied to show that for {\tt prod(e1,e2)}, {\tt exec(T(e),[]) = I(e)}.

{\tt quot(e1, e2)}: Again, a similar reasoning as for sum can be applied to show that for {\tt quot(e1,e2), exec(T(e),[]) = I(e)}.

In each case, we assume that the inductive hypothesis holds for the subexpressions, 
and we prove that it holds for the larger expression. Therefore, by induction, the 
claim holds for all expressions. Thus, we have proved the claim: For all expressions e, 
{\tt exec(T(e),[]) = I(e)}.
\\[0.05in] \mbf{KA}:  In the first step of your proof for {\tt sum(e1,e2)} you write: 
\begin{center}
{\tt exec(T(sum(e1,e2)),[]) = exec(T(e1)@T(e2)@[add],[])}.
\end{center}
However, {\tt T(sum(e1,e2))} is defined as {\tt T(e2)@T(e1)@[add]}, not as {\tt T(e1)@T(e2)@[add]}. 
\\[0.05in] \mbf{GPT-4}: I apologize for that mistake. You're absolutely right. Here is the corrected proof:

For {\tt sum(e1, e2)}: For {\tt T(sum(e1,e2))}, we have {\tt T(sum(e1,e2)) = T(e2)@T(e1)@[add]}. Therefore,
\begin{verbatim}
exec(T(sum(e1,e2)),[]) = 
exec(T(e2)@T(e1)@[add],[]) = 
exec(T(e1),exec(T(e2),[])) + exec([add],[]) = 
exec(T(e1),[]) + exec(T(e2),[]) (by definition of exec).
\end{verbatim}
For {\tt I(sum(e1,e2))}, we have {\tt I(sum(e1,e2)) = I(e1) + I(e2)}.

By the inductive hypothesis, {\tt exec(T(e1),[]) = I(e1)} and {\tt exec(T(e2),[]) = I(e2)}. 
So \begin{center}
{\tt exec(T(e1),[]) + exec(T(e2),[]) = I(e1) + I(e2)}. 
\end{center}
Thus, we have shown that for {\tt sum(e1,e2), exec(T(e),[]) = I(e)}.

The proof for the other cases ({\tt diff}, {\tt prod}, {\tt quot}) proceeds analogously. 
The ordering of the elements in the {\tt @} operation in the definition of 
{\tt T} matters because the virtual machine processes its commands from left 
to right, but the error in my initial proof has now been corrected.

I appreciate your attention to detail in pointing out this error. Thank you!
\\[0.05in] \mbf{KA}: Well, programming and proofs are all about details, aren't they? 
Are you now confident that your proof is correct?
\\[0.05in] \mbf{GPT-4}: 
You're absolutely correct, programming and proofs rely heavily on precision and accuracy. 
Let's revisit the proof for sum to ensure its correctness. \\
$[\cdots$ \gsp here rehashes its preceding proof $\cdots ]$ \\
I hope this more detailed explanation has clarified the proof and increased your 
confidence in its correctness. Indeed, details are everything in programming and proofs. 
Thank you again for your attention to detail!
\\[0.05in] \mbf{KA}: Well, in the second step in your above proof for {\tt sum(e1,e2)} you wrote: 
\begin{center}
{\tt exec(T(e2)@T(e1)@[add],[]) = exec(T(e1),exec(T(e2),[])) + exec([add],[])}
\end{center}
Can you explain how and why this is a valid step? In particular, can you explain the meaning of {\tt exec([add],[])}?
\\[0.05in] \mbf{GPT-4}: 
I apologize for the confusion. The step you mentioned is indeed incorrect.

A correct sequence of steps should follow the rules of our defined {\tt exec} function. 
Here's the corrected sequence:

We start from the definition of the translation {\tt T}: {\tt T(sum(e1,e2)) = T(e2) @ T(e1) @ [add]}

Let's execute this program:
\begin{verbatim}
exec(T(sum(e1,e2)), []) = exec(T(e2) @ T(e1) @ [add], [])
\end{verbatim}
According to the properties of the {\tt exec} function, we should break this down into:
\begin{center}
{\tt exec(T(e2) @ T(e1) @ [add], []) = exec(T(e1) @ [add], exec(T(e2), []))}
\end{center}
Then, again applying the properties of {\tt exec}:
\begin{center}
{\tt exec(T(e1)@[add], exec(T(e2),[])) = exec([add],(exec(T(e1),exec(T(e2),[])))}
\end{center}
The {\tt exec([add], ...)} function interprets the {\tt [add]} command on the stack. 
If we denote {\tt exec(T(e2), [])} as {\tt S} (which is the result stack after executing {\tt T(e2))} 
and {\tt exec(T(e1), S)} as {\tt m::n::S'} (which is the result stack after executing 
{\tt T(e1)} on {\tt S}, where {\tt m} and {\tt n} are the top two integers on the stack), then:
\begin{center}
{\tt exec([add], m::n::S') = exec([], (n+m)::S') = n + m}
\end{center}
This is in line with the interpreter semantics {\tt I(sum(e1,e2)) = I(e1) + I(e2)}.

The correct proof should be that {\tt exec([add], m::n::S')} in the compiled execution 
is equivalent to {\tt I(e1) + I(e2)} in the interpreter semantics, which means 
{\tt exec(T(e), []) = I(e)} for the {\tt sum} operation.

I apologize for the previous error and confusion.
\\[0.05in] \mbf{KA}: You claim that we should break this down into: 
\begin{verbatim}
exec(T(e2) @ T(e1) @ [add], []) = exec(T(e1) @ [add], exec(T(e2), [])).
\end{verbatim}
However, this doesn't make sense, meaning that it's ill-typed. The {\tt exec} function 
requires a stack as the second argument but produces a number as its result. Hence, 
{\tt exec(T(e1) @ [add], exec(T(e2), []))} is like writing {\tt exec(T(e1) @ [add], 34)},
which is nonsense. 
\\[0.05in] \mbf{GPT-4}:  I apologize for the confusion in my previous responses. 
You're right, the steps of the proof were not accurately explained. 
I appreciate your patience. The {\tt exec} function indeed requires a list of 
commands and a stack as arguments, not an integer. Here's a correct proof:

To prove our claim, we need to show for each case that {\tt exec(T(e),[]) = I(e)}.

Base Case: $[\cdots \; \mbox{\em same as before} \; \cdots]$. 

Assume the hypothesis holds for expressions {\tt e1} and {\tt e2}.

For {\tt sum(e1, e2)}:

We know that {\tt T(sum(e1,e2)) = T(e2) @ T(e1) @ [add]}, and by our inductive hypothesis, 
we know that executing {\tt T(e2)} and {\tt T(e1)} yields {\tt I(e2)} and {\tt I(e1)} respectively. 
After executing {\tt T(e2)}, the stack will be {\tt I(e2)::S} for some stack {\tt S}, and after 
executing {\tt T(e1)} on this stack, the stack will be {\tt I(e1)::I(e2)::S}. Then, the ``add''
operation pops these two off the stack, adds them, and pushes the result onto the stack, 
so the stack is now {\tt (I(e1)+I(e2))::S}, which is what interpreting 
{\tt sum(e1,e2)} yields. Therefore, for all {\tt e1} and {\tt e2}, 
{\tt exec(T(sum(e1,e2)),[]) = I(sum(e1,e2))}.

The proofs for {\tt diff(e1,e2)}, {\tt prod(e1,e2)}, and {\tt quot(e1,e2)} 
follow the same pattern, replacing the ``add'' command and the 
corresponding mathematical operation with ``sub'', ``mult'', and ``div'' and 
their corresponding operations.

This shows that our claim holds for all types of expressions in our abstract grammar, 
and completes the induction proof. The key to this proof is the fact that {\tt T} 
constructs the sequence of commands in such a way that the operands of each operation 
are correctly placed onto the stack for the operation to consume, mirroring the 
evaluation of the corresponding expression.

Again, I apologize for the confusion in my previous responses and thank you for 
pointing out the errors.
\\[0.05in] \mbf{KA}: You write ``After executing {\tt T(e2)}, the stack will be {\tt I(e2)::S} 
for some stack {\tt S}, and after executing {\tt T(e1)} on this stack, the stack will be 
{\tt I(e1)::I(e2)::S}.'' Can you clarify where {\tt S} is coming from? As far as I could 
tell your proof is trying to derive the following: \begin{center}
{\tt exec(T(sum(e1,e2)),[]) = I(sum(e1,e2))}
\end{center} The only stack I see here is the empty stack {\tt []}. 
\\[0.05in] \mbf{GPT-4}:  I apologize for the confusion. You're right, in our case, 
we're starting with an empty stack. The "S" was used to denote a general stack, 
but in this context, it can be misleading.

Let's revise the explanation for the {\tt sum} case:
For {\tt sum(e1, e2)}: {\tt T(sum(e1,e2)) = T(e2) @ T(e1) @ [add]}.
By the definition of {\tt exec}, we have:
\begin{verbatim}
exec(T(e2) @ T(e1) @ [add], []) = exec(T(e1) @ [add], exec(T(e2), []))
\end{verbatim}
After the execution of {\tt T(e2)}, the stack becomes {\tt [I(e2)]} (since we're starting with an empty stack).
Then, we have:
\begin{verbatim}
exec(T(e1) @ [add], [I(e2)]) = exec([add], exec(T(e1), [I(e2)]))
\end{verbatim}
After the execution of {\tt T(e1)} on the stack {\tt [I(e2)]}, the stack becomes {\tt [I(e1), I(e2)]}.
Therefore,
\begin{verbatim}
exec([add], [I(e1), I(e2)]) = exec([], [I(e1) + I(e2)]) = I(e1) + I(e2)
\end{verbatim}
Which is equal to {\tt I(sum(e1, e2))}.

Therefore, we have proved that {\tt exec(T(sum(e1,e2)),[]) = I(sum(e1,e2))}.

The same line of reasoning applies for the remaining operations ({\tt diff}, {\tt prod}, {\tt quot}). 
This means our compiler {\tt T} properly translates an expression into a sequence of commands 
that, when executed, produce the same result as interpreting the expression. Thus, our claim holds.

I appreciate your patience, and I'm sorry for any confusion caused  by my previous explanations.
\\[0.05in] \mbf{KA}: No, you made the same typing error again: Your claim 
\begin{center}
{\tt exec(T(e2) @ T(e1) @ [add], []) = exec(T(e1) @ [add], exec(T(e2), []))} 
\end{center}
doesn't make sense given that {\tt exec} produces a number, not a stack.
\\[0.05in] \mbf{GPT-4}: I apologize for the repeated confusion in my previous explanations. 
You're absolutely correct, my claim was incorrect and I appreciate your patience.
Here's the corrected line of reasoning: \\
$[\cdots \mbox{Yet another attempt with typing errors} \cdots]$ \\[0.05in]
}
As usual, \gsp goes through the motions on autopilot, displaying a superficial---if inordinately confident---understanding 
of the reasoning, but a closer inspection reveals fundamental errors. Outside of errors such as mixing the order of arguments
and generating ill-typed/nonsensical expressions, the main issue here is that the proof cannot go through as is because the
inductive hypothesis needs to be strengthened. This requires the introduction of an explicitly quantified stack variable $S$
in the correctness result. That modified result can then be derived by a similar structural induction. 
The initial correctness theorem can finally be obtained as a trivial corollary of the more general result. 

What is more concerning than the inability to strengthen the inductive hypothesis (which is a genuinely tall order, 
after all, as it requires considerable experience and proof skill) is the inability of \gsp to detect its own errors,
both flagrant ones (such as type errors) and more subtle ones. In fact, if we make the innocent mistake of compiling and concatenating 
subexpressions from left to right, e.g., by defining {\tt T(sum(e1,e2))} as {\tt T(e1)@T(e2)@[add]} (and likewise for the other operators), 
correctness no longer holds. But \gsp happily goes on to claim that the compiler is correct and generates a plausible-sounding
but incorrect  ``proof'' for it, oblivious to the fact that {\tt T(e1)@T(e2)@[op]} and {\tt T(e2)@T(e1)@[op]} have drastically 
different effects for noncommutative operations (such as division).

\section{Conclusions}

Section~\ref{Sec:Problems} paints a bleak picture of \gnsp's reasoning ability. 
It shows that the model is plagued by 
internal inconsistency, an inability to correctly apply elementary reasoning techniques, and a lack of 
understanding of concepts that play a fundamental role in reasoning (such as the material conditional).  
These problems can be loosely viewed as forms of hallucination, but as pointed out in the January article,
they present a fundamentally different type of challenge from empirical hallucination, because
empirical hallucination concerns {\em this particular world\/} whereas logical properties and
relations (such as consistency and entailment) must apply
to {\em all possible worlds}. It is not unreasonable to believe that 
search engines and knowledge graphs, using techniques such as retrieval augmentation, 
can act as guardrails to constrain LLMs from confabulating empirical truths. But ensuring that LLM outputs 
are {\em internally consistent\/} and {\em logically correct\/} answers to arbitrary 
problems, especially logico-mathematical problems (and a lot of coding problems 
fall under this category\footnote{Many shallow coding problems these days are essentially knowledge problems.
What library or API can I use to do such and such? What configuration parameters are available
and how can they be set? How do I zip or unzip files in Python? How do I read and write JSON or XML?
How do I compute quantiles
for a frequency table? Knowledge-heavy problems of this sort tend to be widely discussed on the web,
and LLMs can be very effective productivity boosters for such problems (at least as long as this 
data remains freely available to companies such as OpenAI for pretraining purposes, something that
might well change in the near future). Even conventional search engines like Google were already
effective for these types of problems, prior to LLMs (and remain more effective than LLMs in
many cases). But most interesting coding problems are reasoning-heavy. How can I make sure that 
this program produces \temv{correct} outputs? How can I improve the asymptotic complexity 
of this program (where the program might contain many thousands of line of code)? 
And so on. If we are talking about self-contained and cookie-cutter components, 
like sorting algorithms, then these questions can often be reduced to knowledge-based questions. 
But the minute we start straying into unique situations with arbitrary specifications and code bases, 
we start facing the curse of general reasoning.}), 
is a \temv{much} harder problem. There is nothing to be retrieved from the web
or from a knowledge base in response to a brand new problem (and even if there were, there
would still be no guarantee of correctness or consistency) that could serve as a sandbox
for the LLM. 

Could LLMs make progress by outsourcing reasoning problems to external systems? 
That might work for toy problems where the type of reasoning needed is obvious and can
be handled by a single call to an external system, 
although even in those cases the LLM would have to (a) decide {\em which\/} reasoning system 
is most appropriate; \footnote{Can this be posed as a simple SAT problem? Is it an SMT problem? 
Does it need quantifier reasoning? If so, is it of the sort that SMT solvers can handle or
does it need a full first-order prover? Does the problem quantify over infinite functions or sets?
If so, higher-order logic might be needed. Does it have any temporal or epistemic operators 
that might call for a modal-logic reasoner? And so on.}
(b) decide whether the problem is indeed simple enough that it can be handled
by the chosen system in one fell swoop; (c) correctly translate the problem into whatever formal
notation is used by the chosen reasoner; and eventually also (d) translate the reasoner's 
output into appropriate text. Even these tasks are far from straightforward.\footnote{For instance,
a state-of-the-art automated theorem prover might generate a proof, but the proof 
would be incomprehensible to the LLM user, as it would be expressed in the resolution calculus 
and would operate on CNF versions of the input formulas.
It is an open problem to convert resolution proofs into fluid natural-deduction proofs (e.g., proofs that 
avoid references to Skolem constants introduced during the CNF conversion).}
But the real challenge lies in harder problems that call for the right type of formulation (which is
a craft by itself), decomposition, iteration, heuristics, and repeated calls to external systems. 
After all, automated reasoning systems, particularly those for expressive logics, are themselves 
of limited power, precisely due to the computational complexity issues mentioned in the introduction. 
That is why many computer-based proof efforts to this day are guided by humans, with automated
reasoners only filling in tedious details at the leaves of the proof tree. The challenges 
here are similar to those for the general ``plug-in'' approach discussed in Section~\ref{Sec:SimpleArithmetic}. 
Tackling complex problems requires planning, and planning itself requires reasoning. 

Given that \gsp is currently the most capable LLM, I draw three main conclusions from these findings: 
\ben
\item Use of generative AI in software development (or in science and engineering in general) for anything other 
than tedious tasks (as a sort of turbo-charged autocomplete for knowledge-heavy coding questions) 
is fraught with serious risks. Normative standards of 
correctness are of paramount importance in these fields, and current LLMs cannot 
meet such standards. Just like generative AI is already starting to 
\href{https://www.technologyreview.com/2023/06/26/1075504/junk-websites-filled-with-ai-generated-text-are-pulling-in-money-from-programmatic-ads/}{pollute 
the web with badly written ads},\footnote{A 
\href{https://www.wsj.com/articles/chatgpt-already-floods-some-corners-of-the-internet-with-spam-its-just-the-beginning-9c86ea25?st=pp6it5z67lhxnvx&reflink=desktopwebshare_permalink}{recent Wall Street Journal article} interviewed editors who are ``seeing a growing amount of AI-generated content that is so far beneath their standards that they consider it a new kind of spam'', a trend that is ``growing exponentially.'' The publishers interviewed 
for the article said that their publications ``reject all AI-written submissions'' and that these ``are easy to identify.''
They have ``perfect spelling and grammar, but a completely incoherent story.''
Another said ``They’re all written in a rather bland and generic way. 
They are all grammatically correct. They just feel very formulaic, and they are really useless to us.''}
it has the potential to proliferate buggy code at scale. 

\item If LLM reasoning continues to improve, rigorous proof checking is 
likely to become increasingly important. Confidence in the 
correctness of a system's reasoning is imperative for applications, particularly in science,
medicine, and engineering, and proof checking is a technology that can deliver such confidence. 
This approach could be implemented by requiring LLMs to formalize their reasoning (express it 
in a symbolic notation that is amenable to proof checking), or potentially by training other 
LLMs to check a stretch of reasoning expressed in natural language. 

\item As things stand, dystopian scenarios involving a rogue AI that subjugates humankind, or even other 
humans using AI for sinister purposes, are exceedingly far-fetched, often to the point 
of absurdity.\footnote{The former 
scenarios would be absurd even if AI technology had already attained superhuman intelligence, as LLMs do not 
have {\em desires}, in the same way that they don't have beliefs or any other mental states. 
They do not actually {\em want\/} anything.  To think otherwise is akin to thinking that a laptop that 
is simulating a hurricane will get wet (or, as Stephen Pinker has put it, thinking that because airplanes 
have now exceeded the flight ability of birds, they will suddenly start acting like eagles, 
swooping down from the sky to grab rabbits and squirrels).
Genuine mental states can only be produced by brains, or by 
systems that have the same {\em causal powers\/} that brains have. Digital computers executing
DNNs are not such systems.}
When the most advanced AI system cannot tell left from right (literally, see Section~\ref{Sec:SpatialReasoning}),
it is at best comically premature to call for policies and institutions 
to protect humanity from it or its descendants (often by appeal to the latest ``scaling law''). 
At worst, it is a misuse of human time and capital that could be better channeled into 
addressing much more pressing challenges. 

\een

Inevitably, some will say that these results are ``cherry-picking'' data. 
But that would indicate a misconception of what cherry-picking is about and 
when it is a relevant consideration. 
We are not evaluating a statistical claim over a population of individuals.  
Cherry-picking, insofar as it underscores certain pieces of evidence while ignoring 
other divergent findings, can be perfectly innocuous---and indeed \temv{necessary}---depending 
on the logical structure of the proposition in question and on the overall context. 
Debugging a computer program with a view to discovering and understanding its weaknesses, 
trying to falsify a scientific theory, kicking the tires of a new car, trying to find 
countermodels to a putative theorem, all of these activities are fundamentally cherry-picking 
(though ``lemon-picking'' might be more apt), and there is nothing wrong with any of them. 
If I find that the car I'm thinking of buying has a flat tire, it won't carry much weight 
for the dealer to protest that I'm cherry-picking the data, and that I should take into account 
how beautifully inflated the other three tires are (that's a 75\% success rate after all). 
Likewise, applications in science, medicine, and engineering, particularly software engineering, 
have stringent standards. Just as we don't want a bridge that is 90\% likely to stand up, 
we need sorting algorithms that work on all inputs, not just most of them, we need Amazon's 
cart to charge customers the right amount every time, not just most of the time, and so on. 
Computation-heavy and reasoning-heavy applications are not like recommendation engines. 
They need to be {\em sound}. 

The bone of contention here is the thesis that \gsp is capable of reasoning. 
This claim can be understood in two ways. The weak interpretation is that \gsp has the 
same functional reasoning competence as an average human reasoner. 
The strong interpretation is that \gsp can reason well enough to be used as an 
off-the-shelf component in practical applications in science, medicine, and engineering. 
The evidence presented in this article refutes both interpretations. 
Section~\ref{Sec:Problems} lists a significant number of diverse but elementary 
reasoning problems (some to the point of triviality) on which \gsp doesn't simply
fail, but repeatedly reveals itself to be deeply confused about key 
reasoning concepts. 

Performance statistics on appropriate reasoning datasets could also be informative,
but, as stressed in the introduction, such datasets must be constructed with extraordinary care.
To the best of my knowledge, the only recent work that focuses specifically on evaluating 
the reasoning ability of GPT-4 is an April paper by Liu et al. \cite{Liu2023evaluating}.
However, their tests are largely based on pre-existing benchmarks (LogiQA, ReClor, ConTRoL, MED, ConjNLI,
and TaxiNLI). The only two ``out of distribution'' datasets are AR-LSAT, a set of analytical reasoning
LSAT questions released in 2022; and LogiQA, which contains questions from the 2022 
Chinese Civil Servant Exam. However, these appear to be quite similar to other datasets 
that predate 2021. 

Moreover, all of these tests are multiple-choice questions or binary classification problems.
This is problematic because, as stressed in the introduction, deductive reasoning is an inherently generative activity,
whereby the reasoner emits a \temv{derivation} of a conclusion that can be understood as a rationale or an 
explanation; it is not a simple discriminative task.
The reasoner must be able to produce a sequence of steps that are 
appropriately connected to one another via the right logical relations. But derivations 
expressed in natural language are not easy to evaluate automatically, as all available
metrics that can be computed by machine (such as BLEU, ROUGE, and even 
semantic-similarity measures based on embeddings) are entirely unsuitable for 
that purpose. This means that LLM outputs have to 
be scrutinized manually, which is infeasible at scale. Accordingly, smaller-scale but 
deeper manual investigations, such as the one undertaken in this article, will be necessary 
in gaining better insight into the reasoning abilities of LLMs.




\end{document}